\pdfoutput=1

\documentclass[11pt]{article}

\usepackage[final]{acl}

\usepackage{times}
\usepackage{latexsym}

\usepackage[T1]{fontenc}
\usepackage[utf8]{inputenc}

\usepackage{microtype}

\usepackage{inconsolata}

\usepackage{graphicx}
\usepackage{float}

\usepackage{enumitem}
\usepackage{amsmath}
\usepackage{booktabs}
\usepackage{subcaption}
\usepackage{amssymb}
\usepackage{xcolor}
\usepackage{colortbl}
\usepackage{cleveref}

\newcommand{\sminmax}[2]{{\tiny$_{[\text{#1},\text{#2}]}$}}

\title{Beyond Prompts: A Systematic Study of LLM-Based Machine Translation Evaluators}
\hypersetup{
  pdftitle={Beyond Prompts: A Systematic Study of LLM-Based Machine Translation Evaluators},
  pdfauthor={Dmitry Popov, Roman Bokhyan, Ekaterina Enikeeva, Artem Mekhraliev, Stepan Vysotsky, Nikolay Karpachev}
}

\author{
  \textbf{Dmitry Popov}\textsuperscript{\ensuremath{\clubsuit}},
  \textbf{Roman Bokhyan}\textsuperscript{\ensuremath{\diamondsuit}},
  \textbf{Ekaterina Enikeeva}\textsuperscript{\ensuremath{\clubsuit}}, \\
  \textbf{Artem Mekhraliev}\textsuperscript{\ensuremath{\clubsuit}},
  \textbf{Stepan Vysotsky}\textsuperscript{\ensuremath{\clubsuit}},
  \textbf{Nikolay Karpachev}\textsuperscript{\ensuremath{\clubsuit}} \\
  \ensuremath{\clubsuit} Yandex, \quad
  \ensuremath{\diamondsuit} HSE University
}

\begin{document}
\maketitle
\begin{abstract}
LLM-based evaluators have rapidly become a standard tool for
machine translation evaluation, yet the design choices that
shape their behavior remain poorly understood. We present a
systematic study of how instruction design, few-shot
demonstrations, backbone model choice, and multi-step
workflows affect span-level and ranking-based MT evaluation. We
collect new expert RATE annotations on WMT24 and WMT25
English--Russian translations, with transfer validation on MQM-
and ESA-annotated benchmarks across five additional language
pairs. Our experiments reveal four robust findings: instruction
structure shapes evaluator behavior, high-quality demonstrations
matter more than quantity, frontier backbones exhibit distinct
evaluator profiles (recall-, precision-, or ranking-oriented),
and fusing outputs across diverse backbones outperforms
single-evaluator multi-call setups. Although evaluator design
exposes a large combinatorial space, several discovered design
principles transfer consistently across annotation protocols
and language pairs. We synthesize these findings into
\textsc{Orbit}, a family of LLM-based MT evaluators. Our
strongest configurations achieve expert-level performance in error
span detection and in Accuracy and Fluency ranking at roughly
$20\times$ lower cost than expert human annotation.
\end{abstract}

\begin{table}[t]
\centering
\resizebox{\columnwidth}{!}{%
\setlength{\tabcolsep}{4pt}
\begin{tabular}{lccccc c}
\toprule
 & \multicolumn{5}{c}{\textbf{Span-level}} &
   \multicolumn{1}{c}{\textbf{Ranking}} \\
\cmidrule(lr){2-6}\cmidrule(lr){7-7}
\textbf{System} & R & P & F1 & R\textsubscript{4+} & P\textsubscript{4+} & MQM \\
\midrule
Human\textsubscript{E}        & 46.8 & 44.3 & 45.5 & 65.3 & 65.3 & 75.3 \\
\arrayrulecolor{gray!50}\midrule\arrayrulecolor{black}
Human\textsubscript{A-SbS}   & 39.3 & 47.1 & 42.8 & 55.7 & 63.2 & 68.0 \\
Human\textsubscript{A-PW}    & 31.8 & 46.4 & 37.7 & 49.8 & 57.1 & 65.0 \\
Human\textsubscript{WMT}     & 12.3 & 61.9 & 20.6 & 26.6 & 62.8 & 42.7 \\
\midrule
GEMBA-MQM                     & 24.6 & 43.0 & 31.3 & 45.0 & 51.0 & 58.7 \\
GEMBA-ESA                     & 24.9 & 41.3 & 31.0 & 44.6 & 45.7 & 57.3 \\
MQM-AE                        & 24.0 & 51.8 & 32.8 & 45.4 & 54.5 & 60.0 \\
ESA-AE                        & 27.0 & 54.8 & 36.2 & 49.5 & 58.6 & 53.7 \\
UMB-v5                        & 27.8 & 44.5 & 34.2 & 48.7 & 56.4 & 57.3 \\
xCOMET-XXL                    & 18.3 & 36.4 & 24.3 & 33.5 & 36.8 & 58.3 \\
\midrule
\textsc{Orbit-SC}             & 51.5 & 46.2 & 48.7 & 74.7 & 63.5 & 67.0 \\
\textsc{Orbit-MC}             & 60.3 & 45.8 & \textbf{52.1} & \textbf{82.0} & \textbf{70.6} & \textbf{69.3} \\
\bottomrule
\end{tabular}}
\caption{Overview of human baselines, automatic baselines, and
our configurations. Prompt-based baselines and \textsc{Orbit-SC}
use GPT-5.4. \textsc{Orbit-MC} is a multi-call setup that fuses
three evaluator backbones with an LLM fuser.}
\label{tab:main}
\end{table}

\section{Introduction}

Machine translation (MT) quality has improved rapidly in
recent years, with frontier LLM-based systems approaching
professional human translation quality on many language pairs
and domains \citep{kocmi2024findings}. As translation quality
improves, evaluation increasingly becomes the bottleneck:
identifying subtle translation errors requires trained
annotators, substantial annotation time, and careful quality
control. Automatic metrics remain useful at scale, but still lag
behind expert human judgment on fine-grained translation
assessment, particularly for span-level error localization and
diagnostic analysis \citep{lavie2025findings, freitag2021experts}.

Large language models have recently emerged as strong MT
evaluators \citep{kocmi2023gemba}. Prior work has explored both
score-based approaches, which predict a single quality score for
a translation \citep{juraska2024metricx}, and span-based approaches,
which attempt to identify concrete translation errors following
human evaluation protocols such as MQM \citep{freitag2021experts} and
ESA \citep{kocmi2024esa}. However, evaluator design choices ---
prompt structure, demonstrations, workflow decomposition, and aggregation ---
are typically treated as implementation details, despite
substantial evidence from in-context learning research that
demonstration design strongly affects downstream behavior
\citep{liu2022makes, agrawal2023contextual}. PrExMe explores prompt
variation for scalar-score MT evaluation~\citep{leiter2024prexme}.
We instead study
protocol-grounded evaluator design for fine-grained span-level and
ranking-based assessment against expert annotations.

This leaves a central question largely unexplored: which
evaluator design choices robustly improve LLM-based MT
evaluation quality across datasets, protocols, and annotator
settings?

In this work, we study LLM-based MT evaluation as a structured
evaluator design problem rather than as a fixed prompting configuration.
We conduct a systematic study across four major design
dimensions: instruction design, few-shot demonstration design,
backbone model selection, and multi-step evaluator workflows.
Our focus is span-level MT evaluation under the RATE
protocol~\citep{popov2025rate}, which provides fine-grained error
spans, severity annotations, and separate Accuracy and Fluency
judgments.

To support this study, we collect new expert annotations on
English--Russian WMT24 and WMT25 translations under the RATE
protocol. Each translation is independently annotated by
multiple professional linguists, enabling direct evaluation of error
span detection against human annotations rather than relying
only on scalar quality-score correlations. Span-level evaluation is
particularly important for diagnostic MT assessment, since it
localizes concrete translation failures and provides actionable
feedback for both system developers and human editors.

Our experiments reveal four robust findings: instruction
structure shapes evaluator behavior, high-quality demonstrations
matter more than quantity, frontier backbones exhibit distinct
evaluator profiles (recall-, precision-, or ranking-oriented),
and multi-model fusion substantially outperforms single-call
evaluation. Although evaluator design exposes a large
combinatorial space, several discovered design principles
transfer consistently across annotation protocols and language
pairs. Across these experiments, our strongest configurations
consistently exceed assessor-level agreement and match expert-level
error span detection performance (Table~\ref{tab:main}).

We encode these design principles in \textsc{Orbit}, a family of evaluator
configurations instantiating the most robust choices
identified in our study. We refer to the resulting single-call
and multi-call configurations as \textsc{Orbit-SC} and
\textsc{Orbit-MC} respectively. Our strongest configurations achieve
expert-level error span detection performance. \textsc{Orbit-MC} also
reaches expert-level Accuracy and Fluency ranking (\Cref{tab:rq5}) and
exceeds assessor agreement on MQM ranking (\Cref{tab:main}). We
additionally provide a preliminary study of how
expert annotators interact with LLM-generated reports in
collaborative annotation workflows (Appendix~\ref{sec:hmc}).

Our contributions are: (i) A systematic study of evaluator design for
fine-grained LLM-based MT evaluation. (ii) Evidence that several evaluator design principles transfer across annotation protocols (RATE, MQM, ESA), annotator pools, and five additional WMT language directions: en-de, zh-en, cs-de, en-it, and en-zh.
(iii) New expert RATE annotations on WMT24 and WMT25
English--Russian translations spanning multiple annotator expertise
levels.\footnote{\url{https://huggingface.co/datasets/yandex/orbit-mt-eval}}
(iv) \textsc{Orbit}, a family of single- and multi-call MT
evaluators whose strongest configuration achieves expert-level error
span detection performance and surpasses
assessor-level translation ranking while
costing roughly $20\times$ less than expert human annotation.

\section{Background and Related Work}
\label{sec:background}

\subsection{Human MT Evaluation Protocols}

MQM~\citep{lommel2014mqm, freitag2021experts} and
ESA~\citep{kocmi2024esa} serve as de facto gold standards in WMT
evaluation campaigns. Both ask annotators to identify error spans
in translations. MQM additionally assigns each span an error
category and severity, which are aggregated into a final score,
while ESA removes categories and instead uses spans as guidance
for a final 0--100 quality score.

RATE~\citep{popov2025rate} extends this setup with five-point
severity labels, a streamlined semantic/non-semantic taxonomy,
separate Accuracy and Fluency scores, and an overall score
derived from span severities.

\subsection{LLM-Based MT Evaluation}

LLM-based evaluation has emerged as a general paradigm for natural
language assessment~\citep{zheng2023judging, gu2024survey,
liu2023geval, bavaresco2024llms}. In MT evaluation, existing
approaches can be broadly divided into score-based, span-based,
and supervised methods.

Score-based approaches such as GEMBA-DA~\citep{kocmi2023gemba}
and MetricX~\citep{juraska2024metricx} predict a single quality
score for a translation segment. Prompt-based span evaluators
more closely follow human annotation workflows by asking LLMs to
identify translation errors directly. Canonical examples include
GEMBA-MQM~\citep{kocmi2023gembamqm} and
AutoMQM~\citep{fernandes2023devil}, which prompt LLMs using
MQM-style instructions and a small set of demonstrations.
Subsequent work extends this paradigm with decomposition,
post-editing validation, and rubric-based
prompting~\citep{lu2024mqmape, kim2025rubric, lu2024error}.
Multi-agent approaches such as M-MAD~\citep{chen2024mmad}
further decompose evaluation into interacting specialist agents.

PrExMe~\citep{leiter2024prexme} systematically explores more
than 720 prompt templates for open-source LLM-based MT and
summarization metrics, showing that seemingly minor prompt and
output-format choices can alter metric behavior. Our study has a
different focus: we measure direct agreement with multiple
professional experts on error localization and severity, and
jointly study demonstrations, backbone choice, and multi-call
workflows in addition to instruction variation.

Supervised approaches instead train models on MQM-style
annotations to predict scores, spans, or both. For example,
xCOMET~\citep{guerreiro2024xcomet} combines sentence-level
scoring with span detection, while recent work such as
GemSpanEval~\citep{juraska2025metricx} adapts frontier LLMs to
span-level evaluation tasks.

Span-level results are sensitive to annotation quality and annotator
expertise, while systematic comparisons against multiple professional
experts remain limited~\citep{freitag2021experts,lavie2025findings}.

Concurrent work by~\citet{perrella2026spanlevel} proposes
alternative span-matching formulations, which we discuss in
Appendix~\ref{app:metric-robustness}.

\section{Experimental Setup}

\subsection{Design Study Protocol}
\label{sec:design-study}

Evaluator design exposes a large combinatorial space of
potentially interacting decisions, including instruction
structure, few-shot demonstration selection, workflow
decomposition, aggregation, and backbone model choice.
Exhaustive joint optimization across all dimensions is
impractical. We therefore study design dimensions sequentially,
fixing variables after each stage to isolate the effect of subsequent decisions. Our objective is not to find a universally optimal evaluator configuration, but to identify design choices
that consistently improve span-level MT evaluation quality and
transfer across protocols and language pairs.

\subsection{Dataset and Annotation Setup}
\label{sec:dataset}

We collect new annotations on 600 source--translation pairs
from WMT24~\cite{kocmi2024findings} and WMT25~\cite{kocmi2025findings} English--Russian translations, extending
the RATE-annotated coverage to newer
frontier LLM translation systems. Detailed dataset statistics
are in Appendix~\ref{app:dataset}.

\paragraph{Sources and translations.}
Source segments are sampled from WMT24 and WMT25 across multiple
domains. We select 300 source segments and pair each with two
translations, enabling segment-level pairwise ranking accuracy
(\Cref{sec:metrics}) alongside span-level evaluation. Rather
than sampling uniformly, we apply diversity sampling to ensure
coverage across translation systems and quality levels,
producing a challenging evaluation set.

\paragraph{Annotation layers and human baselines.}
Each translation receives annotation at three levels of expertise.
Professional expert linguists annotate the dataset under the full
RATE protocol in implicit side-by-side mode, yielding three
independent expert annotations per segment
(\textbf{Expert~$\times$~3}). Assessors --- Russian-native
annotators with verified C1 English proficiency but no formal
translation training --- follow the same protocol
(\textbf{Assessor}). We additionally include standard ESA
annotations from the WMT evaluation campaign
(\textbf{WMT annotator}), providing a link to the public
benchmark setting.

For evaluation, we keep two of the three expert annotations as
gold references. Human\textsubscript{E} represents pairwise
expert agreement, computed by evaluating the third expert against
each gold annotation and averaging --- the same procedure used
for all evaluated systems.
Human\textsubscript{A-SbS} and
Human\textsubscript{A-PW} correspond to assessors evaluated in
implicit side-by-side and point-wise setups respectively, while
Human\textsubscript{WMT} follows standard ESA annotation
conditions. Full annotator criteria and statistics are in
Appendix~\ref{app:annotators}.

\paragraph{Few-shot pool.}
Separately from the evaluation set, we construct a dedicated
few-shot calibration pool consisting of 8 source segments across
multiple domains, each translated by 12 systems. Professional
linguists annotate the pool collaboratively under the RATE
protocol with consensus reconciliation, yielding more reliable
demonstrations than single-annotator outputs. This pool is used
exclusively for few-shot selection (\Cref{sec:rq2}) and does not
overlap with the evaluation set.

\paragraph{Multilingual validation.}
To test whether discovered evaluator design choices transfer
beyond English--Russian and the RATE protocol, we additionally
evaluate our best single-call configuration on five WMT language
directions with publicly available annotations. From the WMT23
Metrics Shared Task~\citep{freitag2023results}, we use MQM
annotations on en-de and zh-en, sampling 450 examples per
direction. From the WMT25 evaluation shared
task~\citep{lavie2025findings}, we use ESA annotations on cs-de,
en-it, and en-zh, sampling 300 examples per direction. Full
results are in \Cref{sec:multilingual}.

\subsection{LLM Baselines}
\label{sec:baselines}

We compare against baselines representing several major evaluator
paradigms: concise prompting, detailed protocol prompting,
decomposition-based workflows, and supervised span-level
evaluation.

Prompt-based baselines include
GEMBA-MQM~\citep{kocmi2023gembamqm},
GEMBA-ESA, a reference-free LLM judge used in the WMT25 General MT
shared task~\citep{kocmi2025findings}, and UMB-v5, our abbreviation
for the \texttt{unified-mqm-boosted-v5} schema released with
\citet{perrella2026spanlevel}. The latter is a detailed prompt that
thoroughly describes the annotation task. We additionally include
MQM-AE and ESA-AE as our own adaptations of MQM and ESA human
annotation guidelines into LLM prompts, keeping deviations from
the originals minimal. As a supervised baseline we include
xCOMET-XXL~\citep{guerreiro2024xcomet}, trained on MQM-style
annotations to predict both scores and spans. All evaluation prompts
are available in our
\href{https://huggingface.co/datasets/yandex/orbit-mt-eval}{Hugging Face release}.
Omitted methods are discussed in Appendix~\ref{app:methods-not-included}.

\subsection{Evaluation Metrics}
\label{sec:metrics}

We evaluate MT evaluators along two dimensions: span-level error
detection and segment-level pairwise ranking accuracy. Throughout
the paper, MQM scores are derived from RATE annotations following
\citet{popov2025rate}: RATE severity levels 2--3 map to MQM minor
errors and levels 4--5 to major errors. As in the original
RATE-to-MQM conversion, RATE severity~1 (``negligible or arguable''
errors) is excluded because standard MQM contains no equivalent
severity level.

\paragraph{Span-level metrics.}
Predicted error spans are matched to gold annotations via
bipartite matching with character overlap (MP with $\tau{=}1$).
We report micro-averaged recall $R$, precision $P$, and harmonic
mean $F_1$, following~\citet{perrella2026spanlevel}. Segment-level
macro-averaging gives equal weight to each segment regardless of its
number of annotated errors and can therefore overstate performance
for low-recall evaluators. To evaluate severe-error
behavior, we additionally report two complementary diagnostics:
$R_{4+}$, recall over gold spans of severity~$\geq 4$ regardless
of predicted severity, and $P_{4+}$, precision for predictions
assigned severity~$\geq 4$ against gold major-error spans. For
MQM- and ESA-annotated data,
$R_{4+}$ and $P_{4+}$ are computed over major and critical
errors. We verify robustness under alternative span-matching
formulations in Appendix~\ref{app:metric-robustness}.

\paragraph{Ranking accuracy.}
For each pair of translations of the same source, we measure
whether the evaluator correctly predicts which translation is
better across Accuracy (Acc), Fluency (FL), and MQM dimensions.

\begin{table}[ht]
\centering
\resizebox{\columnwidth}{!}{%
\setlength{\tabcolsep}{4pt}
\begin{tabular}{lccccc ccc}
\toprule
 & \multicolumn{5}{c}{\textbf{Span-level}} & \multicolumn{3}{c}{\textbf{Ranking}} \\
\cmidrule(lr){2-6}\cmidrule(lr){7-9}
\textbf{System} & R & P & F1 & R\textsubscript{4+} & P\textsubscript{4+} & MQM & Acc & FL \\
\midrule
Human\textsubscript{E}    & 46.8 & 44.3 & 45.5 & 65.3 & 65.3 & 75.3 & 67.3 & 70.0 \\
\midrule
GEMBA-MQM                 & 19.9 & 42.1 & 27.0 & 39.6 & 51.1 & 60.7 & --   & --   \\
GEMBA-ESA                 & 20.8 & 39.6 & 27.3 & 39.3 & 53.0 & 60.7 & --   & --   \\
MQM-AE                    & 22.2 & 53.8 & 31.4 & 42.9 & 64.4 & 63.0 & --   & --   \\
ESA-AE                    & 26.3 & 54.5 & 35.5 & 47.5 & 61.9 & 59.3 & --   & --   \\
UMB-v5                    & 22.0 & 47.0 & 30.0 & 42.7 & 57.9 & 55.0 & --   & --   \\
\midrule
RATE-AE\textsubscript{brief}    & 26.5 & 48.3 & 34.2 & 46.3 & 58.6 & 59.7 & 61.7 & 64.3 \\
RATE-AE\textsubscript{full}     & 23.9 & 44.7 & 31.2 & 42.2 & 55.2 & 58.3 & 61.5 & 64.9 \\
RATE-AE\textsubscript{full+ex}  & 30.9 & 53.8 & 39.3 & 52.9 & 66.3 & 56.7 & 60.3 & 65.3 \\
\bottomrule
\end{tabular}}
\caption{Effect of instruction design (zero-shot, GPT-5.2 backbone). GEMBA systems use their original few-shot demonstrations. All other systems are zero-shot.}
\label{tab:rq1}
\end{table}

\section{Systematic Study of LLM Evaluation Design}

\subsection{Instruction Design}
\label{sec:rq1}

How closely should evaluator instructions mirror human
annotation protocols? Existing prompt-based evaluators adopt
substantially different assumptions: GEMBA-MQM and AutoMQM rely
on concise language-agnostic prompts, while UMB-v5 follows the
opposite strategy with a detailed protocol-oriented
specification. Neither line of work ablates the instruction
content itself, leaving the relationship between protocol detail
and evaluator behavior unmeasured.

We compare three families of protocol-adapted prompts (MQM-AE,
ESA-AE, RATE-AE), with RATE-AE ablated into three variants of
decreasing detail: RATE-AE\textsubscript{full+ex} contains full
category definitions with inline error-span examples (as
category-boundary disambiguators, not few-shot demonstrations),
RATE-AE\textsubscript{full} keeps definitions and severity
anchors without examples, and RATE-AE\textsubscript{brief}
compresses to a minimal task description, severity scale, and
output format. All experiments in this and the following
section (\Cref{sec:rq2}) use GPT-5.2 with temperature 0.

\Cref{tab:rq1} shows large differences across instruction
variants, with span F1 ranging from 27.0 (GEMBA-MQM) to 39.3
(RATE-AE\textsubscript{full+ex}). MQM- and ESA-style prompts
achieve high precision but low recall, including on
high-severity errors --- indicating genuinely missed problems
rather than annotation granularity differences. GEMBA systems
exhibit similar behavior despite using few-shot demonstrations.
The strongest RATE variant improves span F1 and major-error
recovery over the protocol baselines in this GPT-5.2 comparison.
Appendix~\ref{app:backbone-replication} further shows that all
tested RATE variants outperform the protocol baselines on F1 for
Claude Opus~4.6 and Gemini~3 Flash. This effect is not
specific to RATE-annotated gold: RATE-AE\textsubscript{brief}
also outperforms MQM-AE on F1 and $R_{4+}$ when evaluated against
MQM-annotated WMT data (\Cref{sec:multilingual}).

Within the RATE family, the relationship between instruction
detail and evaluator quality is non-monotonic: removing the
inline examples is costly, with RATE-AE\textsubscript{full}
falling behind even the much shorter
RATE-AE\textsubscript{brief}. This indicates that detailed
taxonomy definitions alone overconstrain evaluator behavior,
while compact instructions preserving only task-critical
guidance generalize more effectively. At the same time, the
strong performance of RATE-AE\textsubscript{full+ex} indicates
that detailed protocols remain useful when augmented with
concrete examples, foreshadowing the demonstration-based
calibration we study in \Cref{sec:rq2}. Ranking accuracy
remains stable across variants, indicating that segment-level
judgments are less sensitive to instruction structure than to
the backbone (\Cref{sec:rq3}).

Overall, human annotation guidelines transfer effectively to
LLM evaluators, but exhaustive taxonomy descriptions become
less important once high-quality demonstrations or compact task
specifications are available. We use
RATE-AE\textsubscript{full+ex} as our primary zero-shot setup
and revisit this choice in \Cref{sec:rq2}.

\begin{table}[thbp]
\centering
\resizebox{\columnwidth}{!}{%
\setlength{\tabcolsep}{4pt}
\begin{tabular}{lccc ccc}
\toprule
 & \multicolumn{3}{c}{\textbf{Span-level}} & \multicolumn{3}{c}{\textbf{Ranking}} \\
\cmidrule(lr){2-4}\cmidrule(lr){5-7}
\textbf{Few-shot source} & F1 & R\textsubscript{4+} & P\textsubscript{4+} & MQM & Acc & FL \\
\midrule
RATE-AE\textsubscript{full+ex} (0-shot) & 39.3 & 52.9 & 66.3 & 56.7 & 60.3 & 65.3 \\
\midrule
WMT        & 37.0\sminmax{36.4}{37.7} & 50.1\sminmax{49.8}{50.4} & 69.6\sminmax{69.2}{70.0} & 61.1\sminmax{60.7}{62.0} & 64.5\sminmax{63.2}{66.0} & 61.5\sminmax{60.7}{62.7} \\
RATE-WMT24 & 38.8\sminmax{37.5}{40.3} & 52.8\sminmax{51.0}{55.1} & 67.9\sminmax{66.3}{70.3} & 61.6\sminmax{61.0}{62.3} & 64.1\sminmax{63.0}{65.3} & 63.9\sminmax{63.7}{64.0} \\
Ours       & \textbf{41.2}\sminmax{40.9}{41.7} & \textbf{56.8}\sminmax{56.0}{57.5} & 68.1\sminmax{67.8}{68.7} & \textbf{62.4}\sminmax{59.0}{67.3} & 63.4\sminmax{63.0}{64.3} & \textbf{65.3}\sminmax{64.3}{67.0} \\
\bottomrule
\end{tabular}}
\caption{\small Effect of few-shot annotation quality. Each value
is the mean over 3 runs. Brackets show the corresponding
minimum and maximum.}
\label{tab:rq2-quality}
\end{table}

\subsection{Few-Shot Design}
\label{sec:rq2}

\subsubsection{Does Annotation Quality Matter?}

We compare three sources of few-shot demonstrations: WMT25
shared task ESA annotations (manually mapped to
the RATE taxonomy and severity scale by our expert annotators),
single-expert RATE annotations on WMT24
from~\citet{popov2025rate}, and our
collaboratively curated annotations produced by multiple experts
with consensus reconciliation.

\Cref{tab:rq2-quality} shows that demonstration quality matters
substantially more than demonstration presence alone. WMT
annotations push F1 below the zero-shot baseline across all
three runs (37.0 vs 39.3 mean), suggesting that noisy
demonstrations can actively mislead the evaluator about what
constitutes a translation error. RATE-WMT24 broadly matches the
zero-shot baseline on mean F1. Only our collaboratively curated
annotations consistently improve both overall span recovery and
major-error detection.

In contrast, ranking metrics improve over the zero-shot baseline
more uniformly across demonstration sources, indicating that span-level evaluator
behavior is substantially more affected by demonstration design
than pairwise ranking judgments. We therefore fix our curated
annotation pool for all subsequent few-shot experiments.

\begin{figure}[htbp]
\centering
\includegraphics[width=\columnwidth]{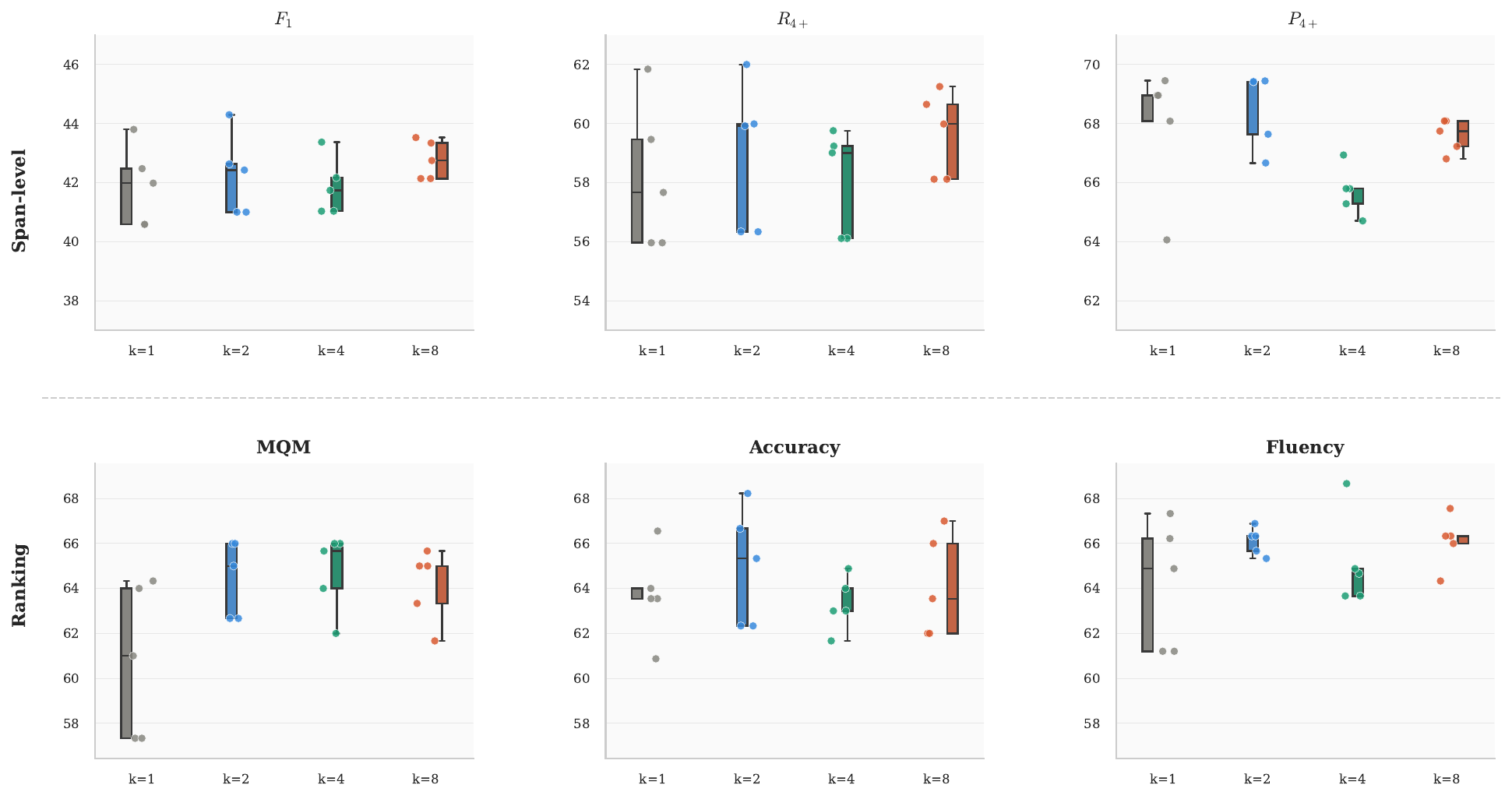}
\caption{Effect of few-shot example count ($k$) across span-level 
and ranking metrics. Each condition uses 5 random draws. 
Boxes show IQR, points show individual runs.}
\label{fig:rq2-quantity}
\end{figure}

\subsubsection{How Many Examples?}

With the demonstration source fixed, we next vary the number of
examples ($k \in \{1,2,4,8\}$, \Cref{fig:rq2-quantity}).

Increasing $k$ produces only modest changes in mean performance.
Instead, the primary effect of larger few-shot sets is variance
reduction: $k{=}1$ exhibits noticeably less stable behavior,
while $k \geq 2$ yields broadly consistent results across
independently sampled configurations. The most consistent mean
trend is a slight improvement in R\textsubscript{4+}, suggesting
that additional demonstrations primarily improve major-error
recovery rather than general error detection. We adopt
$k{=}2$ for the remaining experiments, revisiting selection
strategy in \Cref{sec:rq2-3}.

\subsubsection{Coverage-Based Selection}
\label{sec:rq2-3}

The previous experiments showed that evaluator behavior depends
strongly on demonstration design, motivating a more structured
selection strategy. We construct compact few-shot sets that span
the protocol's error categories and severity range. We evaluate
$k \in \{2,3,4\}$, corresponding to approximately 85\%, 96\%,
and 100\% category/severity coverage.

\begin{table}[ht]
\centering
\resizebox{\columnwidth}{!}{%
\setlength{\tabcolsep}{4pt}
\begin{tabular}{lccccc ccc}
\toprule
 & \multicolumn{5}{c}{\textbf{Span-level}} & \multicolumn{3}{c}{\textbf{Ranking}} \\
\cmidrule(lr){2-6}\cmidrule(lr){7-9}
\textbf{Few-shot config} & R & P & F1 & R\textsubscript{4+} & P\textsubscript{4+} & MQM & Acc & FL \\
\midrule
RATE-AE\textsubscript{full+ex} (0-shot)          & 30.9 & 53.8 & 39.3 & 52.9 & 66.3 & 56.7 & 60.3 & 65.3 \\
\midrule
\multicolumn{9}{l}{\texttt{Coverage-based selection (varying $k$), with RATE-AE\textsubscript{full+ex}}} \\
$k{=}2$                                          & 39.8 & 51.1 & 44.8 & 63.4 & 67.8 & 64.0 & 62.7 & 69.3 \\
$k{=}3$                                          & 35.8 & 53.4 & 42.9 & 59.8 & 66.5 & 60.3 & 62.6 & 64.6 \\
$k{=}4$                                          & 37.8 & 53.2 & 44.2 & 61.6 & 68.1 & 63.7 & 65.7 & 65.3 \\
\midrule
\multicolumn{9}{l}{\texttt{Coverage-based $k{=}2$, varying instruction}} \\
$\hookrightarrow$ RATE-AE\textsubscript{full+ex} & 39.8 & 51.1 & 44.8 & 63.4 & 67.8 & 64.0 & 62.7 & 69.3 \\
$\hookrightarrow$ RATE-AE\textsubscript{full}    & 39.5 & 52.2 & 44.9 & 63.6 & 66.6 & 63.3 & 64.7 & 69.3 \\
$\hookrightarrow$ RATE-AE\textsubscript{brief} (\textsc{Orbit-SC})  & 42.4 & 49.6 & 45.7 & 64.8 & 67.5 & 64.0 & 65.3 & 64.3 \\
$\hookrightarrow$ GEMBA-MQM                      & 31.0 & 39.4 & 34.7 & 49.5 & 49.1 & 65.3 & --   & --   \\
$\hookrightarrow$ GEMBA-ESA                      & 31.8 & 37.0 & 34.2 & 51.5 & 50.0 & 63.3 & --   & --   \\
\bottomrule
\end{tabular}}
\caption{\small Effect of coverage-based few-shot selection.
Upper block: varying $k$ with RATE-AE\textsubscript{full+ex}.
Lower block: $k{=}2$ with varying instruction.}
\label{tab:rq2-src}
\end{table}

\Cref{tab:rq2-src} shows that compact coverage-based selection
already substantially improves span-level quality over the
zero-shot setup. Performance is broadly comparable across $k$,
with a mild recall--precision trade-off: smaller $k$ favors
recall, while larger $k$ improves severity-aware precision at
the cost of recall. This is consistent with the earlier
observation that demonstration quantity has limited effect on
mean performance, primarily reducing run-to-run variance.

The coverage criterion itself is defined without optimizing
metric scores. We select $k{=}2$ as the smallest configuration
with comparable validation performance, balancing quality,
prompt complexity, and inference cost.

Revisiting instruction choice under this setup reverses the
zero-shot result from \Cref{sec:rq1}:
RATE-AE\textsubscript{brief} now slightly surpasses
RATE-AE\textsubscript{full+ex} on span F1, suggesting that
high-quality few-shot demonstrations partially substitute for
detailed taxonomy descriptions. Instruction design nevertheless
remains important: replacing the RATE-style prompts with
GEMBA-MQM or GEMBA-ESA substantially reduces span-level quality
even under the same few-shot configuration.

We refer to this single-call configuration ---
RATE-AE\textsubscript{brief} paired with coverage-based $k{=}2$
demonstrations --- as \textsc{Orbit-SC}. \Cref{sec:rq3}
evaluates this configuration across a range of frontier LLM
backbones.

\begin{table}[htbp]
\centering
\resizebox{\columnwidth}{!}{%
\setlength{\tabcolsep}{4pt}
\begin{tabular}{l ccccc ccc}
\toprule
 & \multicolumn{5}{c}{\textbf{Span-level}} & \multicolumn{3}{c}{\textbf{Ranking}} \\
\cmidrule(lr){2-6}\cmidrule(lr){7-9}
\textbf{Model} & R & P & F1 & R\textsubscript{4+} & P\textsubscript{4+} & MQM & Acc & FL \\
\midrule
Human\textsubscript{E}     & 46.8 & 44.3 & 45.5 & 65.3 & 65.3 & 75.3 & 67.3 & 70.0 \\
\midrule 
GPT-4.1                    & \textbf{51.7} & 45.6 & \textbf{48.5} & \underline{72.7} & 61.7 & 61.0 & 59.0 & 57.0 \\
GPT-5.2                    & 42.4 & 49.9 & 45.9 & 64.7 & 65.4 & 65.0 & 62.3 & 66.0 \\
GPT-5.4                    & \textbf{51.5} & 46.2 & \textbf{48.7} & \textbf{74.7} & 63.5 & 67.0 & 65.0 & 65.0 \\
Claude Sonnet 4.6          & 45.8 & 45.6 & 45.7 & 67.1 & 63.5 & 64.0 & 59.5 & 62.9 \\
Claude Opus 4.6            & \underline{47.8} & 46.9 & \underline{47.3} & \underline{72.2} & \textbf{68.3} & 65.7 & \textbf{68.3} & \textbf{70.7} \\
Gemini 3 Flash             & 38.9 & 49.1 & 43.4 & 62.2 & 66.8 & 68.0 & 59.3 & 67.0 \\
Gemini 3.1 Pro             & 45.0 & \textbf{51.2} & \underline{47.9} & 67.8 & 66.9 & \textbf{74.7} & 67.1 & 68.5 \\
Grok-4                     & \underline{50.4} & 35.8 & 41.8 & 71.0 & 52.5 & 56.7 & 55.9 & 49.8 \\
DeepSeek-V3.2              & 41.8 & 17.4 & 24.6 & 60.7 & 15.3 & 53.0 & 54.1 & 52.7 \\
DeepSeek-R1                & 39.6 & 35.0 & 37.2 & 59.2 & 46.6 & 53.3 & 60.5 & 55.9 \\
Kimi K2 Thinking           & 40.9 & 42.1 & 41.5 & 59.5 & 50.4 & 57.7 & 57.2 & 63.2 \\
Qwen-235B                  & 38.9 & 39.4 & 39.2 & 59.3 & 50.7 & 52.7 & 53.5 & 46.8 \\
\bottomrule
\end{tabular}}
\caption{\small Model comparison using \textsc{Orbit-SC} across frontier models.}
\label{tab:rq3}
\end{table}

\subsection{Model Comparison}
\label{sec:rq3}

How do frontier LLM backbones compare as MT evaluators? Using
the \textsc{Orbit-SC} configuration, we evaluate a range of
frontier models as evaluator backbones. \Cref{tab:rq3} reports
the main results. Extended metrics are provided in
Appendix~\ref{app:rq3-extended}.

Different frontier models exhibit distinct evaluator profiles,
with no single backbone dominating all evaluation dimensions.
Recent GPT models are recall-oriented: GPT-5.4 achieves the
strongest span-level recovery and major-error detection, while
GPT-4.1 provides similar recall behavior at substantially lower
cost. Claude Opus 4.6 exhibits the strongest precision-oriented
profile, leading on severity-aware precision and
direct-assessment ranking while remaining competitive on span F1.
Gemini 3.1 Pro combines ranking and precision strengths: it
achieves the strongest MQM ranking accuracy together with the
highest overall precision among span-level evaluators.

At the other end of the spectrum, Grok-4 and DeepSeek-V3.2
substantially over-predict errors at a large precision cost, with
DeepSeek-V3.2 in
particular flagging more than twice as many errors per segment
as Human\textsubscript{E}. The remaining open-weight models
(DeepSeek-R1, Kimi K2 variants, Qwen-235B) sit consistently
below the top tier across both span-level and ranking-based
evaluation.

Overall, strong evaluator quality is multi-dimensional:
different models favor different trade-offs between recall,
precision, severity calibration, and ranking behavior. We adopt
GPT-5.4 as the default backbone in the remaining experiments.
Appendix~\ref{app:backbone-replication} repeats the matched design
comparison on Claude Opus~4.6 and Gemini~3 Flash to test whether
the broader design conclusions transfer beyond GPT-5.4.

\subsection{Point-Wise vs. Side-by-Side Evaluation}
\label{sec:sbs}

Human MT evaluation is commonly conducted in an implicit side-by-side
(SbS) setting, where one annotator sees multiple translations of the
same source~\citep{freitag2021experts}. Our main \textsc{Orbit-SC} setup
is point-wise (PW): each call evaluates one source--translation pair
independently. We adapt it to SbS by placing two translations of the
same source in one call and returning a separate report for each. The
few-shot sources remain fixed, with the second translation added to
each demonstration.

\Cref{tab:sbs} shows a clear trade-off. SbS improves Accuracy and
Fluency ranking for all three backbones and improves or matches MQM
ranking, but does not improve span $F_1$. Paired context shifts
predictions toward higher precision and lower recall. PW is therefore
convenient for evaluating an individual MT system, whereas SbS is
better suited to ranking a fixed set of systems. Reversing the
presentation order produced no statistically significant differences.

\begin{table}[H]
\centering
\scriptsize
\setlength{\tabcolsep}{2.4pt}
\resizebox{\columnwidth}{!}{%
\begin{tabular}{ll ccccc ccc}
\toprule
 & & \multicolumn{5}{c}{\textbf{Span-level}} &
 \multicolumn{3}{c}{\textbf{Ranking}} \\
\cmidrule(lr){3-7}\cmidrule(lr){8-10}
\textbf{Backbone} & \textbf{Format} & R & P & F1 &
R\textsubscript{4+} & P\textsubscript{4+} & MQM & Acc & FL \\
\midrule
GPT-5.4          & PW  & 51.5 & 46.2 & 48.7 & 74.7 & 63.5 & 67.0 & 65.0 & 65.0 \\
                 & SbS & 48.7 & 48.8 & 48.8 & 73.5 & 64.2 & 68.3 & 66.7 & 70.3 \\
\midrule
Claude Opus 4.6 & PW  & 47.8 & 46.9 & 47.3 & 72.2 & 68.3 & 65.7 & 68.3 & 70.7 \\
                 & SbS & 42.1 & 50.5 & 45.9 & 66.8 & 71.9 & 65.7 & 71.0 & 73.3 \\
\midrule
Gemini 3 Flash   & PW  & 38.9 & 49.1 & 43.4 & 62.2 & 66.8 & 68.0 & 59.3 & 67.0 \\
                 & SbS & 32.5 & 53.7 & 40.5 & 56.4 & 70.8 & 73.3 & 67.0 & 73.9 \\
\bottomrule
\end{tabular}}
\caption{Point-wise (PW) \textsc{Orbit-SC} and its side-by-side
(SbS) adaptation.}
\label{tab:sbs}
\end{table}

\subsection{Multi-Call Workflows}
\label{sec:rq4}

We next study multi-call workflows as a design dimension: do
additional LLM calls improve evaluator quality? We compare
three approaches: task decomposition, iterative self-review,
and fusion across multiple evaluators (\Cref{tab:rq5}).

\subsubsection{Task decomposition}
Inspired by M-MAD-style decomposition~\citep{chen2024mmad}, we split annotation into
semantic and non-semantic passes. Concatenating both outputs
improves recall, particularly for major errors, but substantially
hurts precision due to overlapping and conflicting annotations.
LLM fusion recovers precision at the cost of recall. In our
setting, decomposition alone provides limited benefit over the
single-call baseline.

\subsubsection{Self-review}
We add a second-pass reviewer that re-examines the first-pass
output, either only adding missed spans or also editing existing
predictions. Self-review consistently improves recall, especially
on major errors, but also introduces borderline predictions that
reduce precision. Allowing the reviewer to edit existing spans
partially recovers this precision loss and yields a small F1 gain
over the single-call setup.

\subsubsection{Model-diversity fusion}
Fusion across diverse evaluator backbones produces the strongest
overall gains. We refer to this multi-call configuration ---
fusing Claude Opus 4.6, GPT-5.4, and Gemini 3.1 Pro outputs via
Claude Opus 4.6 as fuser --- as \textsc{Orbit-MC}.
\textsc{Orbit-MC} matches or exceeds expert-to-expert agreement on span F1
and major-error recovery, while maintaining strong precision on
high-severity errors.

Unlike fixed-label ensembles, these structured reports do not
provide a canonical unit for majority voting: models may identify
the same error with different span boundaries and descriptions,
or assign conflicting categories, severities, and segment scores.
A deterministic rule would therefore first require a separate
span-matching policy and arbitrary conflict-resolution choices.
The semantic fuser is used precisely to perform this reconciliation.

Fusion quality depends strongly on the fuser backbone itself.
Using the same three constituent models, GPT-5.4 as fuser falls
several F1 points below Claude, while Gemini 3.1 Pro as fuser
produces the strongest MQM ranking accuracy. Claude substantially
outperforms GPT-5.4 as a fusion model despite weaker standalone
recall (\Cref{sec:rq3}), suggesting that strong evaluator
generation and strong evaluator aggregation are distinct
capabilities.

\begin{table}[ht]
\centering
\resizebox{\columnwidth}{!}{%
\setlength{\tabcolsep}{4pt}
\begin{tabular}{lccccc ccc}
\toprule
 & \multicolumn{5}{c}{\textbf{Span-level}} & \multicolumn{3}{c}{\textbf{Ranking}} \\
\cmidrule(lr){2-6}\cmidrule(lr){7-9}
\textbf{System} & R & P & F1 & R\textsubscript{4+} & P\textsubscript{4+} & MQM & Acc & FL \\
\midrule
Human\textsubscript{E}                & 46.8 & 44.3 & 45.5 & 65.3 & 65.3 & 75.3 & 67.3 & 70.0 \\
\midrule
\multicolumn{9}{l}{\texttt{Single-call}} \\
\textsc{Orbit-SC} (GPT-5.4)           & 51.5 & 46.2 & 48.7 & 74.7 & 63.5 & 67.0 & 65.0 & 65.0 \\
\midrule
\multicolumn{9}{l}{\texttt{Multi-call: dimension split}} \\
Sem/non-sem concatenated              & 54.8 & 37.6 & 44.6 & 76.9 & 52.0 & 69.7 & 64.1 & 63.7 \\
Sem/non-sem fused                     & 41.0 & 51.1 & 45.5 & 64.1 & 64.9 & 63.3 & 62.3 & 64.0 \\
\midrule
\multicolumn{9}{l}{\texttt{Multi-call: self-review}} \\
Add missing spans                     & 58.4 & 40.5 & 47.8 & 80.3 & 59.4 & 65.7 & 63.0 & 67.0 \\
Add + edit spans                      & 52.2 & 46.3 & 49.1 & 76.5 & 62.1 & 65.0 & 62.3 & 65.0 \\
\midrule
\multicolumn{9}{l}{\texttt{Multi-call: model diversity fusion (C+G+Gem, varying fuser)}} \\
$\hookrightarrow$ Claude Opus 4.6 fuser ($\textsc{Orbit-MC}$)  & \textbf{60.3} & 45.8 & \textbf{52.1} & \textbf{82.0} & \textbf{70.6} & 69.3 & \textbf{67.7} & \textbf{73.7} \\
$\hookrightarrow$ GPT-5.4 fuser                                & 48.0 & 48.0 & 48.0 & 68.5 & 66.0 & 64.0 & 66.7 & 70.7 \\
$\hookrightarrow$ Gemini 3.1 Pro fuser                         & 51.7 & 49.7 & 50.7 & 74.1 & 71.6 & \textbf{70.0} & 65.9 & 71.1 \\
\midrule
\multicolumn{9}{l}{\texttt{Multi-call: varying constituent set (Claude Opus 4.6 fuser)}} \\
2 models (C+G)                        & 55.4 & 46.5 & 50.5 & 78.6 & 69.3 & 68.3 & 70.3 & 69.3 \\
5 models (C+G+Gem+Grok+Kimi)          & \textbf{62.6} & 42.1 & 50.3 & \textbf{84.3} & 66.2 & 67.3 & 70.3 & 72.3 \\
\midrule
\multicolumn{9}{l}{\texttt{Multi-call: other fusion strategies (Claude Opus 4.6 fuser)}} \\
Temperature diversity (3 runs)        & 57.0 & 45.9 & 50.8 & 78.7 & 66.1 & 68.3 & 65.3 & 70.0 \\
Strategy diversity                    & 56.0 & 45.1 & 50.0 & 78.7 & 67.1 & 68.0 & 65.0 & 69.3 \\
\bottomrule
\end{tabular}}
\caption{\small Multi-call strategies. C\,=\,Claude Opus 4.6, G\,=\,GPT-5.4, Gem\,=\,Gemini 3.1 Pro. Fused\,=\,LLM-merged report.}
\label{tab:rq5}
\end{table}

\begin{table}[t]
\centering
\small
\setlength{\tabcolsep}{4pt}
\begin{tabular}{llr}
\toprule
\textbf{Evaluator setup} & \textbf{Format} & \textbf{USD/10k seg.} \\
\midrule
\multicolumn{3}{l}{\textit{Human, one annotator pass}} \\
Professional expert RATE & SbS & \$48,920 \\
Assessor RATE            & SbS & \$6,500 \\
Assessor RATE            & PW  & \$6,865 \\
\midrule
\multicolumn{3}{l}{\texttt{ORBIT-SC, varying backbone and format}} \\
Gemini 3 Flash   & PW  & \$20 \\
GPT-4.1          & PW  & \$60 \\
GPT-5.4          & PW  & \$85 \\
GPT-5.4          & SbS & \$75 \\
Claude Opus 4.6 & PW  & \$345 \\
Gemini 3.1 Pro  & PW  & \$385 \\
\midrule
\textsc{Orbit-MC} (3 models + fuser) & PW & \$1,720 \\
\bottomrule
\end{tabular}
\caption{API cost and estimated cost of one human annotator pass per
10,000 evaluated segments.}
\label{tab:cost}
\end{table}

\begin{table*}[!t]
\centering
\resizebox{\textwidth}{!}{%
\setlength{\tabcolsep}{4pt}
\begin{tabular}{l ccccc ccccc ccccc ccccc ccccc}
\toprule
 & \multicolumn{5}{c}{\textbf{en-de}\textsuperscript{MQM}}
 & \multicolumn{5}{c}{\textbf{zh-en}\textsuperscript{MQM}}
 & \multicolumn{5}{c}{\textbf{cs-de}\textsuperscript{ESA}}
 & \multicolumn{5}{c}{\textbf{en-it}\textsuperscript{ESA}}
 & \multicolumn{5}{c}{\textbf{en-zh}\textsuperscript{ESA}} \\
\cmidrule(lr){2-6}\cmidrule(lr){7-11}\cmidrule(lr){12-16}\cmidrule(lr){17-21}\cmidrule(lr){22-26}
\textbf{System}
 & R & P & F1 & R\textsubscript{4+} & MQM
 & R & P & F1 & R\textsubscript{4+} & MQM
 & R & P & F1 & R\textsubscript{4+} & MQM
 & R & P & F1 & R\textsubscript{4+} & MQM
 & R & P & F1 & R\textsubscript{4+} & MQM \\
\midrule
Human\textsubscript{2}                       & 31.8 & 40.7 & 35.7 & 48.7 & 60.7 & 44.5 & 41.5 & 42.9 & 63.2 & \textbf{68.7} & 21.8 & 36.0 & 27.2 & 29.3 & 45.3 & 25.5 & 20.3 & 22.6 & 37.2 & 48.0 & 15.5 & 18.1 & 16.7 & 17.8 & 41.3 \\
\midrule
GEMBA-MQM                                    & 29.9 & 35.5 & 32.5 & 43.2 & 60.0 & 29.3 & 44.3 & 35.2 & 46.6 & 63.8 & 17.0 & 22.4 & 19.3 & 23.8 & 46.7 & 17.8 & 10.5 & 13.2 & 29.8 & 43.3 & 33.2 & 12.8 & 18.5 & 43.9 & \textbf{50.7} \\
MQM-AE                                       & 30.1 & 45.5 & 36.2 & 46.2 & 56.9 & 31.6 & 51.7 & 39.3 & 52.0 & 58.0 & 29.1 & 29.5 & 29.3 & 38.4 & 32.0 & 43.8 & 14.7 & 22.0 & 60.3 & 50.0 & 38.8 & 14.4 & 21.0 & 46.7 & 42.0 \\
ESA-AE                                       & 29.8 & 46.8 & 36.4 & 45.2 & 59.8 & 33.2 & 60.6 & 42.9 & 55.7 & 62.2 & 37.7 & 27.9 & \textbf{32.1} & 45.4 & 52.0 & 45.1 & 15.1 & \textbf{22.7} & 63.6 & 46.0 & 42.1 & 14.6 & \textbf{21.7} & 46.7 & 36.0 \\
UMB-v5                                       & 32.5 & 42.6 & 36.9 & 48.0 & 61.3 & 37.7 & 49.1 & 42.6 & 58.3 & 63.1 & 23.5 & 28.9 & 25.9 & 32.7 & 50.0 & 39.9 & 13.6 & 20.3 & 60.3 & 50.0 & 35.5 & 13.1 & 19.1 & 44.9 & 43.3 \\
\midrule
RATE-AE\textsubscript{brief}                 & 35.4 & 43.2 & 38.9 & 51.3 & 61.8 & 41.8 & 55.7 & 47.7 & 62.6 & 64.9 & 32.1 & 27.6 & 29.7 & 41.4 & \textbf{52.0} & 51.0 & 14.2 & 22.2 & 66.9 & 52.7 & 49.9 & 12.9 & 20.4 & 54.2 & 44.0 \\
RATE-AE\textsubscript{brief} + DS            & 38.0 & 40.3 & 39.1 & 53.8 & 57.1 & 47.8 & 49.0 & 48.4 & 67.1 & 65.1 & --   & --   & --   & --   & --   & --   & --   & --   & --   & --   & --   & --   & --   & --   & --   \\
\textsc{Orbit-SC}                            & 41.5 & 37.1 & 39.2 & 58.3 & 60.7 & 50.6 & 49.5 & \textbf{50.0} & \textbf{69.0} & 66.9 & 39.6 & 22.1 & 28.3 & 49.0 & 49.3 & 60.1 & 11.6 & 19.4 & 74.4 & \textbf{53.3} & 61.5 & 10.7 & 18.3 & 58.9 & 40.0 \\
\textsc{Orbit-MC}                            & \textbf{45.2} & 35.5 & \textbf{39.7} & \textbf{62.1} & \textbf{63.8} & \textbf{51.7} & 48.4 & \textbf{49.9} & 68.0 & 67.3 & \textbf{45.3} & 22.9 & 30.4 & \textbf{54.1} & 48.0 & \textbf{70.1} & 10.5 & 18.2 & \textbf{78.5} & 49.3 & \textbf{64.5} &  9.2 & 16.1 & \textbf{59.8} & 45.3 \\
\bottomrule
\end{tabular}}
\caption{Multilingual transfer on WMT23 MQM (en-de, zh-en) and
WMT25 ESA (cs-de, en-it, en-zh) data. DS = direction-specific demonstrations: two few-shot examples
in the target translation direction.}
\label{tab:multilingual}
\end{table*}

\subsubsection{Sources of diversity}
Within model-diversity fusion, expanding from two to three
constituent models improves overall quality, while expanding the
fusion set to five models by adding recall-oriented backbones
(Grok-4, Kimi-K2) further increases recall but reduces precision
and F1, consistent with the over-prediction tendencies observed
in \Cref{sec:rq3}.

Temperature and strategy diversity also improve over the
single-call baseline but produce consistently smaller gains than
model diversity, indicating that complementary evaluator
profiles are more valuable than repeated sampling from the same
backbone.

Overall, multi-call workflows benefit most from genuine evaluator
diversity rather than from iteration or decomposition over a
single backbone. The multi-call configuration \textsc{Orbit-MC}
therefore adopts three-model fusion (Claude Opus 4.6, GPT-5.4,
Gemini 3.1 Pro) via Claude as fuser, achieving the strongest
span-detection profile across all tested configurations.

\subsection{Cost and Deployment Trade-offs}
\label{sec:cost}

\Cref{tab:cost} compares human and API costs at a common scale of
10,000 evaluated segments. We use this volume as a
representative data-collection campaign for training or adapting an
automatic evaluator rather than a one-off evaluation set. A segment is
one source--translation pair. PW is the standard \textsc{Orbit-SC} setup,
with one pair per call. Its SbS adaptation places two translations of the
same source in one call and returns a separate report for each. The human
PW and SbS conditions use the same visibility distinction. Thus, 300 SbS
tasks and 600 PW tasks both yield 600 segment-level reports.

Human estimates use observed aggregate labor for one annotator pass
rather than extrapolating median completion times. Scaling these totals
to 10,000 reports gives roughly 3,542 expert-hours in SbS, 1,625
assessor-hours in SbS, and 1,717 assessor-hours in PW. Monetary estimates
are rounded to the nearest \$5. The \textsc{Orbit-MC} total includes three
evaluators and the fuser.

At this scale, one pass by a professional expert costs about \$48,920. An
assessor pass costs \$6,500 in SbS and \$6,865 in PW. \textsc{Orbit-MC}
costs roughly \$1,720, corresponding to a $4\times$ cost reduction from
the least expensive assessor setup and a $28\times$ reduction from
professional expert RATE. Point-wise \textsc{Orbit-SC} costs
\$20--\$385 depending on the backbone, a further $4$--$86\times$
reduction relative to \textsc{Orbit-MC} while retaining the strong
results in \Cref{tab:rq3}. For GPT-5.4, SbS costs roughly \$75 compared
with \$85 for separate PW calls. The corresponding quality comparison
is reported in \Cref{sec:sbs}.

These dollar totals still understate the practical gap. Human evaluation
requires recruiting and screening annotators who are native speakers of
the target language and have C1/C2 proficiency in the source language.
For expert pools, formal training or a degree in linguistics or
translation is also preferred. Protocol training, calibration,
scheduling, and ongoing quality control add further
overhead~\citep{freitag2021experts,kocmi2024esa,popov2025rate}. Much of
this infrastructure must be established again for each language
direction, which delays evaluation and makes procedures harder to
standardize across pools. API-based evaluation can reuse a common
workflow across directions, although model quality may still vary across
languages and domains.

\section{Multilingual Transfer}
\label{sec:multilingual}

We test whether the evaluator design choices identified on
English--Russian transfer across annotation protocols,
annotation densities, and language pairs. We evaluate
\textsc{Orbit-SC} and \textsc{Orbit-MC} on WMT23 MQM (en-de,
zh-en) and WMT25 ESA (cs-de, en-it, en-zh) benchmarks using
the same inference-time configurations developed on RATE
English--Russian data. Human\textsubscript{2} denotes a second
annotator scored against the gold annotation.

Across both MQM and ESA benchmarks (\Cref{tab:multilingual}),
the main findings of our design study transfer consistently
beyond the English--Russian setting. \textsc{Orbit-SC} remains
competitive across all evaluated language pairs, while
\textsc{Orbit-MC} preserves its strong recall-oriented profile,
improving major-error recovery over \textsc{Orbit-SC} on en-de
and remaining comparable on zh-en.

Compact RATE-style prompting remains competitive on MQM.
Replacing the en-ru few-shot pool with direction-specific
demonstrations (DS) provides no substantial gain over
\textsc{Orbit-SC} on either MQM direction.

Notably, the en-ru pool matches direction-specific demonstrations
on en-de span $F_1$ while improving $R_{4+}$ and MQM, and improves
all three metrics on zh-en. These results suggest that the advantage
comes not from Russian itself but from how the pool was constructed.
The pool covers diverse error categories and severities, with
annotations collaboratively reconciled by multiple experts.

Transfer to ESA benchmarks exhibits substantially lower
precision across all evaluators, likely reflecting the much
sparser annotation density of ESA compared to MQM and RATE.
On span $F_1$, \textsc{Orbit-MC} provides smaller gains than
on MQM and is slightly below \textsc{Orbit-SC} on en-it and
en-zh, reflecting that multi-model fusion expands prediction
sets in a regime where additional spans are heavily penalized.
Nevertheless, transfer gains remain clearest on major-error
recovery, where \textsc{Orbit-MC} consistently exceeds
Human\textsubscript{2} across all ESA directions.

\section{Conclusion}

We systematically study how instruction design, demonstrations,
backbone choice, and multi-call workflows shape LLM-based MT
evaluation across RATE, MQM, ESA, and multiple language pairs.
Curated, collaboratively reconciled demonstrations can match or
outperform larger pools, model families show distinct evaluator
profiles, and fusion yields the strongest overall results.
\textsc{Orbit} instantiates these findings as single- and
multi-call evaluators. Its strongest configurations achieve
expert-level error span detection and Accuracy/Fluency ranking,
while \textsc{Orbit-MC} exceeds assessor agreement on MQM ranking.

\section*{Limitations}

\paragraph{Language coverage.}
Our main design study is conducted on a single high-resource
language pair (English--Russian). We validate the resulting
\textsc{Orbit-SC} and \textsc{Orbit-MC} configurations on five
additional WMT directions (en-de and zh-en for MQM, and cs-de,
en-it, and en-zh for ESA), but coverage of low-resource pairs,
non-Latin scripts beyond Russian and Chinese, and language
families outside Indo-European and Sino-Tibetan remains limited.
Findings on instruction language and few-shot example selection
in particular may require adaptation for typologically distant
target languages.

\paragraph{Closed-source dependencies.}
Our strongest configurations (\textsc{Orbit-SC} with GPT-5.4,
\textsc{Orbit-MC} fusing Claude Opus~4.6, GPT-5.4, and
Gemini~3.1 Pro) depend on closed-source, API-only models whose
weights and training procedures are not public. This affects
reproducibility (exact behaviour may shift across silent model
updates) and longevity (specific model snapshots may be
deprecated). We mitigate the reproducibility concern
through the broad cross-model comparison in \Cref{sec:rq3},
showing that qualitative conclusions hold across model families.

\paragraph{Span-matching methodology.}
Span-level $F_1$ depends on the choice of span-matching
formulation. We use MP with $\tau{=}1$ as the primary formulation
because our main question is whether the evaluator identifies the
annotated error location. MPP assigns partial credit in proportion
to character overlap and is therefore more sensitive to
span-boundary agreement. Appendix~\ref{app:metric-robustness}
reports MPP and LLM-based reconciliation as complementary
sensitivity analyses.

\paragraph{Severity calibration.}
Appendix~\ref{app:rq3-extended} shows that severity assignment is a
distinct capability from error detection, with both human
experts and LLM evaluators substantially under-rating major
errors when stricter severity thresholds are required. We
report detection-oriented $R_{4+}/P_{4+}$ as primary metrics in
the main paper because they isolate the more fundamental
question of error localization, but downstream use cases that
depend on accurate severity assignment (e.g., quality estimation
thresholds, automatic gating) require additional severity
calibration that is not the focus of this work.

\paragraph{Human--LLM collaboration study.}
Our human--LLM collaboration experiments
(Appendix~\ref{sec:hmc}) are exploratory in scope: they measure
annotation revision rather than ground-truth correctness, use
only \textsc{Orbit-MC} as the reference configuration, and were
conducted with a limited annotator group on a single language
pair. Whether the observed shifts in annotation behavior
correspond to improved annotation quality, and how the effects
generalize across configurations and annotator pools, remains
to be studied.

\paragraph{Subjectivity of MT evaluation.}
Translation quality assessment remains partly subjective even
among trained experts. The agreement between our expert
annotators (\Cref{tab:main}) provides a human-agreement reference
point for how closely any evaluator --- human or LLM-based --- can be
expected to match a single ``ground truth''. We treat expert
agreement as the headline target rather than a hard ceiling, and
recognize that ranking differences within this band may not
reflect meaningful quality differences.

\section*{Ethics Statement}

\paragraph{Annotator compensation and consent.}
RATE annotators received compensation at standard
commercial translation rates for their region, with expert
annotators earning approximately double the hourly wage
of assessors, reflecting the higher qualifications required.
No personal data was collected during the annotation
process, and all source segments were screened for
potentially disturbing material prior to annotation.
All annotators were informed about how their annotations
would be used — including for the development and
evaluation of LLM-based MT evaluators — prior to
beginning the tasks.

\paragraph{Human--LLM collaboration study.}
The collaborative annotation study in Appendix~\ref{sec:hmc}
followed the same compensation and consent procedures.
Annotators were informed that they would be reviewing
LLM-generated reports as part of their workflow, and that
their pre- and post-hint annotations would be compared.
No identifying information about annotators is reported
in our analyses.

\paragraph{Annotator feedback.}
Following the annotation, we administered a feedback
questionnaire. The vast majority of annotators reported
a positive experience and found the instructions clear.

\paragraph{Use of closed-source models.}
Our strongest configurations rely on closed-source LLMs
accessed via commercial APIs. We discuss reproducibility
and longevity implications of this choice in the
Limitations section.

\bibliography{custom}

\begin{thebibliography}{26}
\providecommand{\natexlab}[1]{#1}

\bibitem[{Agrawal et~al.(2023)Agrawal, Zhou, Lewis, Zettlemoyer, and
  Ghazvininejad}]{agrawal2023contextual}
Sweta Agrawal, Chunting Zhou, Mike Lewis, Luke Zettlemoyer, and Marjan
  Ghazvininejad. 2023.
\newblock In-context examples selection for machine translation.
\newblock In \emph{Findings of the Association for Computational Linguistics:
  ACL 2023}.

\bibitem[{Bavaresco et~al.(2025)Bavaresco, Bernardi, Bertolazzi, Elliott,
  Fern{\'a}ndez, Gatt, Ghaleb, Giulianelli, Hanna, Koller, Martins, Mondorf,
  Neplenbroek, Pezzelle, Plank, Schlangen, Suglia, Surikuchi, Takmaz, and
  Testoni}]{bavaresco2024llms}
Anna Bavaresco, Raffaella Bernardi, Leonardo Bertolazzi, Desmond Elliott,
  Raquel Fern{\'a}ndez, Albert Gatt, Esam Ghaleb, Mario Giulianelli, Michael
  Hanna, Alexander Koller, Andr{\'e} F.~T. Martins, Philipp Mondorf, Vera
  Neplenbroek, Sandro Pezzelle, Barbara Plank, David Schlangen, Alessandro
  Suglia, Aditya~K Surikuchi, Ece Takmaz, and Alberto Testoni. 2025.
\newblock \href {https://doi.org/10.18653/v1/2025.acl-short.20} {{LLM}s instead
  of human judges? a large scale empirical study across 20 {NLP} evaluation
  tasks}.
\newblock In \emph{Proceedings of the 63rd Annual Meeting of the Association
  for Computational Linguistics (Volume 2: Short Papers)}, pages 238--255,
  Vienna, Austria. Association for Computational Linguistics.

\bibitem[{Feng et~al.(2025)Feng, Su, Zheng, Ren, Zhang, Wu, Wang, and
  Liu}]{chen2024mmad}
Zhaopeng Feng, Jiayuan Su, Jiamei Zheng, Jiahan Ren, Yan Zhang, Jian Wu,
  Hongwei Wang, and Zuozhu Liu. 2025.
\newblock {M-MAD}: Multidimensional multi-agent debate for advanced machine
  translation evaluation.
\newblock In \emph{Proceedings of the 63rd Annual Meeting of the Association
  for Computational Linguistics (Volume 1: Long Papers)}. Association for
  Computational Linguistics.

\bibitem[{Fernandes et~al.(2023)Fernandes, Deutsch, Finkelstein, Riley,
  Martins, Neubig, Garg, Clark, Freitag, and Firat}]{fernandes2023devil}
Patrick Fernandes, Daniel Deutsch, Mara Finkelstein, Parker Riley, Andr{\'e}
  F.~T. Martins, Graham Neubig, Ankush Garg, Jonathan~H. Clark, Markus Freitag,
  and Orhan Firat. 2023.
\newblock The devil is in the errors: Leveraging large language models for
  fine-grained machine translation evaluation.
\newblock In \emph{Proceedings of the Eighth Conference on Machine Translation
  (WMT)}, Singapore. Association for Computational Linguistics.

\bibitem[{Freitag et~al.(2021)Freitag, Foster, Grangier, Ratnakar, Tan, and
  Macherey}]{freitag2021experts}
Markus Freitag, George Foster, David Grangier, Viresh Ratnakar, Qijun Tan, and
  Wolfgang Macherey. 2021.
\newblock Experts, errors, and context: A large-scale study of human evaluation
  for machine translation.
\newblock \emph{Transactions of the Association for Computational Linguistics},
  9:1460--1474.

\bibitem[{Freitag et~al.(2023)Freitag, Mathur, Lo, Avramidis, Rei, Thompson,
  Kocmi, Blain, Deutsch, Stewart, Zerva, Castilho, Lavie, and
  Foster}]{freitag2023results}
Markus Freitag, Nitika Mathur, Chi{-}kiu Lo, Eleftherios Avramidis, Ricardo
  Rei, Brian Thompson, Tom Kocmi, Fr{\'e}d{\'e}ric Blain, Daniel Deutsch, Craig
  Stewart, Chrysoula Zerva, Sheila Castilho, Alon Lavie, and George Foster.
  2023.
\newblock Results of {WMT23} metrics shared task: Metrics might be guilty but
  references are not innocent.
\newblock In \emph{Proceedings of the Eighth Conference on Machine Translation
  (WMT)}, pages 578--628, Singapore. Association for Computational Linguistics.

\bibitem[{Gu et~al.(2026)Gu, Jiang, Shi, Tan, Zhai, Xu, Li, Shen, Ma, Liu,
  Wang, Zhang, Lin, Zhang, Ni, Gao, Wang, and Guo}]{gu2024survey}
Jiawei Gu, Xuhui Jiang, Zhichao Shi, Hexiang Tan, Xuehao Zhai, Chengjin Xu, Wei
  Li, Yinghan Shen, Shengjie Ma, Honghao Liu, Saizhuo Wang, Kun Zhang, Zhouchi
  Lin, Bowen Zhang, Lionel Ni, Wen Gao, Yuanzhuo Wang, and Jian Guo. 2026.
\newblock \href {https://doi.org/10.1016/j.xinn.2025.101253} {A survey on
  {LLM}-as-a-judge}.
\newblock \emph{The Innovation}, 7(6):101253.

\bibitem[{Guerreiro et~al.(2024)Guerreiro, Rei, van Stigt, Coheur, Colombo, and
  Martins}]{guerreiro2024xcomet}
Nuno~M. Guerreiro, Ricardo Rei, Daan van Stigt, Luisa Coheur, Pierre Colombo,
  and Andr{\'e} F.~T. Martins. 2024.
\newblock {xCOMET}: Transparent machine translation evaluation through
  fine-grained error detection.
\newblock \emph{Transactions of the Association for Computational Linguistics
  (TACL)}.

\bibitem[{Juraska et~al.(2024)Juraska, Deutsch, Finkelstein, and
  Freitag}]{juraska2024metricx}
Juraj Juraska, Daniel Deutsch, Mara Finkelstein, and Markus Freitag. 2024.
\newblock {MetricX-24}: The {Google} submission to the {WMT} 2024 metrics
  shared task.
\newblock In \emph{Proceedings of the Ninth Conference on Machine Translation
  (WMT)}.

\bibitem[{Juraska et~al.(2025)Juraska, Domhan, Finkelstein, Nakagawa, Kovacs,
  Deutsch, Wang, and Freitag}]{juraska2025metricx}
Juraj Juraska, Tobias Domhan, Mara Finkelstein, Tetsuji Nakagawa, Geza Kovacs,
  Daniel Deutsch, Pidong Wang, and Markus Freitag. 2025.
\newblock {MetricX-25} and {GemSpanEval}: {Google} {T}ranslate submissions to
  the {WMT25} evaluation shared task.
\newblock In \emph{Proceedings of the Tenth Conference on Machine Translation
  (WMT)}, Suzhou, China. Association for Computational Linguistics.

\bibitem[{Kim(2025)}]{kim2025rubric}
Ahrii Kim. 2025.
\newblock {RUBRIC-MQM}: Span-level {LLM}-as-judge in machine translation for
  high-end models.
\newblock In \emph{Proceedings of the 63rd Annual Meeting of the Association
  for Computational Linguistics (Volume 6: Industry Track)}, pages 147--165,
  Vienna, Austria. Association for Computational Linguistics.

\bibitem[{Kocmi et~al.(2025)Kocmi, Artemova, Avramidis, Bawden, Bojar, Dranch,
  Dvorkovich, Dukanov, Fishel, Freitag, Gowda, Grundkiewicz, Haddow, Karpinska,
  Koehn, Lakougna, Lundin, Monz, Murray, Nagata, Perrella, Proietti, Popel,
  Popovi{\'c}, Riley, Shmatova, Steingr{\'i}msson, Yankovskaya, and
  Zouhar}]{kocmi2025findings}
Tom Kocmi, Ekaterina Artemova, Eleftherios Avramidis, Rachel Bawden,
  Ond{\v{r}}ej Bojar, Konstantin Dranch, Anton Dvorkovich, Sergey Dukanov, Mark
  Fishel, Markus Freitag, Thamme Gowda, Roman Grundkiewicz, Barry Haddow,
  Marzena Karpinska, Philipp Koehn, Howard Lakougna, Jessica~M. Lundin,
  Christof Monz, Kenton Murray, and 10 others. 2025.
\newblock \href {https://aclanthology.org/2025.wmt-1.22/} {Findings of the
  {WMT}25 general machine translation shared task: Time to stop evaluating on
  easy test sets}.
\newblock In \emph{Proceedings of the Tenth Conference on Machine Translation
  (WMT)}, Suzhou, China. Association for Computational Linguistics.

\bibitem[{Kocmi et~al.(2024{\natexlab{a}})Kocmi, Avramidis, Bawden, Bojar,
  Dvorkovich, Federmann, Fishel, Freitag, Gowda, Grundkiewicz, Haddow,
  Karpinska, Koehn, Marie, Monz, Murray, Nagata, Popel, Popovi{\'c}, Shmatova,
  Steingr{\'i}msson, and Zouhar}]{kocmi2024findings}
Tom Kocmi, Eleftherios Avramidis, Rachel Bawden, Ond{\v{r}}ej Bojar, Anton
  Dvorkovich, Christian Federmann, Mark Fishel, Markus Freitag, Thamme Gowda,
  Roman Grundkiewicz, Barry Haddow, Marzena Karpinska, Philipp Koehn, Benjamin
  Marie, Christof Monz, Kenton Murray, Masaaki Nagata, Martin Popel, Maja
  Popovi{\'c}, and 3 others. 2024{\natexlab{a}}.
\newblock Findings of the {WMT24} general machine translation shared task: The
  {LLM} era is here but {MT} is not solved yet.
\newblock In \emph{Proceedings of the Ninth Conference on Machine Translation
  (WMT)}, pages 1--46, Miami, Florida, USA. Association for Computational
  Linguistics.

\bibitem[{Kocmi and Federmann(2023{\natexlab{a}})}]{kocmi2023gembamqm}
Tom Kocmi and Christian Federmann. 2023{\natexlab{a}}.
\newblock {GEMBA-MQM}: Detecting translation quality error spans with {GPT-4}.
\newblock In \emph{Proceedings of the Eighth Conference on Machine Translation
  (WMT)}, pages 768--775, Singapore. Association for Computational Linguistics.

\bibitem[{Kocmi and Federmann(2023{\natexlab{b}})}]{kocmi2023gemba}
Tom Kocmi and Christian Federmann. 2023{\natexlab{b}}.
\newblock Large language models are state-of-the-art evaluators of translation
  quality.
\newblock In \emph{Proceedings of the 24th Annual Conference of the European
  Association for Machine Translation (EAMT)}, pages 193--203, Tampere,
  Finland.

\bibitem[{Kocmi et~al.(2024{\natexlab{b}})Kocmi, Zouhar, Avramidis,
  Grundkiewicz, Karpinska, Popovi{\'c}, Sachan, and Shmatova}]{kocmi2024esa}
Tom Kocmi, Vil{\'e}m Zouhar, Eleftherios Avramidis, Roman Grundkiewicz, Marzena
  Karpinska, Maja Popovi{\'c}, Mrinmaya Sachan, and Mariya Shmatova.
  2024{\natexlab{b}}.
\newblock Error span annotation: A balanced approach for human evaluation of
  machine translation.
\newblock In \emph{Proceedings of the Ninth Conference on Machine Translation
  (WMT)}, Miami, Florida, USA. Association for Computational Linguistics.

\bibitem[{Lavie et~al.(2025)Lavie, Hanneman, Agrawal, Kanojia, Lo, Zouhar,
  Blain, Zerva, Avramidis, Deoghare, Sindhujan, Wang, Adelani, Thompson, Kocmi,
  Freitag, and Deutsch}]{lavie2025findings}
Alon Lavie, Greg Hanneman, Sweta Agrawal, Diptesh Kanojia, Chi-Kiu Lo,
  Vil{\'e}m Zouhar, Frederic Blain, Chrysoula Zerva, Eleftherios Avramidis,
  Sourabh Deoghare, Archchana Sindhujan, Jiayi Wang, David~Ifeoluwa Adelani,
  Brian Thompson, Tom Kocmi, Markus Freitag, and Daniel Deutsch. 2025.
\newblock \href {https://doi.org/10.18653/v1/2025.wmt-1.24} {Findings of the
  {WMT}25 shared task on automated translation evaluation systems: Linguistic
  diversity is challenging and references still help}.
\newblock In \emph{Proceedings of the Tenth Conference on Machine Translation},
  pages 436--483, Suzhou, China. Association for Computational Linguistics.

\bibitem[{Leiter and Eger(2024)}]{leiter2024prexme}
Christoph Leiter and Steffen Eger. 2024.
\newblock \href {https://doi.org/10.18653/v1/2024.emnlp-main.641} {{PrExMe!}
  large scale prompt exploration of open source {LLM}s for machine translation
  and summarization evaluation}.
\newblock In \emph{Proceedings of the 2024 Conference on Empirical Methods in
  Natural Language Processing}, pages 11481--11506, Miami, Florida, USA.
  Association for Computational Linguistics.

\bibitem[{Liu et~al.(2022)Liu, Shen, Zhang, Dolan, Carin, and
  Chen}]{liu2022makes}
Jiachang Liu, Dinghan Shen, Yizhe Zhang, Bill Dolan, Lawrence Carin, and Weizhu
  Chen. 2022.
\newblock What makes good in-context examples for {GPT-3}?
\newblock In \emph{Proceedings of Deep Learning Inside Out (DeeLIO 2022): The
  3rd Workshop on Knowledge Extraction and Integration for Deep Learning
  Architectures}.

\bibitem[{Liu et~al.(2023)Liu, Iter, Xu, Wang, Xu, and Zhu}]{liu2023geval}
Yang Liu, Dan Iter, Yichong Xu, Shuohang Wang, Ruochen Xu, and Chenguang Zhu.
  2023.
\newblock {G-Eval}: {NLG} evaluation using {GPT-4} with better human alignment.
\newblock In \emph{Proceedings of the 2023 Conference on Empirical Methods in
  Natural Language Processing (EMNLP)}, pages 2511--2522, Singapore.
  Association for Computational Linguistics.

\bibitem[{Lommel et~al.(2014)Lommel, Uszkoreit, and Burchardt}]{lommel2014mqm}
Arle Lommel, Hans Uszkoreit, and Aljoscha Burchardt. 2014.
\newblock \href {https://doi.org/10.5565/rev/tradumatica.77} {Multidimensional
  quality metrics ({MQM}): A framework for declaring and describing translation
  quality metrics}.
\newblock \emph{Tradum{\`a}tica}, pages 455--463.
\newblock No. 12.

\bibitem[{Lu et~al.(2025)Lu, Ding, Zhang, Zhang, and Tao}]{lu2024mqmape}
Qingyu Lu, Liang Ding, Kanjian Zhang, Jinxia Zhang, and Dacheng Tao. 2025.
\newblock {MQM-APE}: Toward high-quality error annotation predictors with
  automatic post-editing in {LLM} translation evaluators.
\newblock In \emph{Proceedings of the 31st International Conference on
  Computational Linguistics (COLING)}. Association for Computational
  Linguistics.

\bibitem[{Lu et~al.(2024)Lu, Qiu, Ding, Zhang, Kocmi, and Tao}]{lu2024error}
Qingyu Lu, Baopu Qiu, Liang Ding, Kanjian Zhang, Tom Kocmi, and Dacheng Tao.
  2024.
\newblock Error analysis prompting enables human-like translation evaluation in
  large language models.
\newblock In \emph{Findings of the Association for Computational Linguistics:
  ACL 2024}, pages 8801--8816. Association for Computational Linguistics.

\bibitem[{Perrella et~al.(2026)Perrella, Morales~Agostinho, and
  Zaragoza}]{perrella2026spanlevel}
Stefano Perrella, Eric Morales~Agostinho, and Hugo Zaragoza. 2026.
\newblock Span-level machine translation meta-evaluation.
\newblock \emph{arXiv preprint arXiv:2603.19921}.

\bibitem[{Popov et~al.(2025)Popov, Negodin, Enikeeva, Matrosova, Karpachev, and
  Ryabinin}]{popov2025rate}
Dmitry Popov, Vladislav Negodin, Ekaterina Enikeeva, Iana Matrosova, Nikolay
  Karpachev, and Max Ryabinin. 2025.
\newblock Refined assessment for translation evaluation: Rethinking machine
  translation evaluation in the era of human-level systems.
\newblock In \emph{Findings of the Association for Computational Linguistics:
  EMNLP 2025}, pages 22079--22095, Suzhou, China. Association for Computational
  Linguistics.

\bibitem[{Zheng et~al.(2023)Zheng, Chiang, Sheng, Zhuang, Wu, Zhuang, Lin, Li,
  Li, Xing, Zhang, Gonzalez, and Stoica}]{zheng2023judging}
Lianmin Zheng, Wei-Lin Chiang, Ying Sheng, Siyuan Zhuang, Zhanghao Wu, Yonghao
  Zhuang, Zi~Lin, Zhuohan Li, Dacheng Li, Eric~P. Xing, Hao Zhang, Joseph~E.
  Gonzalez, and Ion Stoica. 2023.
\newblock Judging {LLM}-as-a-judge with {MT}-bench and chatbot arena.
\newblock In \emph{Advances in Neural Information Processing Systems
  (NeurIPS)}.

\end{thebibliography}

\appendix
\section{Dataset Statistics}
\label{app:dataset}

The design study dataset comprises 600 source--translation pairs
constructed from 300 English source segments sampled from
WMT24--WMT25 English--Russian shared task data, each paired with two system
translations. Source selection prioritizes challenging segments
where candidate translations disagree, ensuring evaluators must
discriminate between subtle quality differences rather than
identify obvious failures.

The few-shot pool uses separately selected sources and does not
overlap with this 600-pair evaluation set.

\paragraph{Domain coverage.}
The dataset covers four domains
(\Cref{tab:dataset-domains}), with literary, news, and speech
segments slightly over-represented relative to social media
text.

\begin{table}[ht]
\centering
\small
\begin{tabular}{lr}
\toprule
\textbf{Domain} & \textbf{Segments} \\
\midrule
Literary & 166 \\
News     & 146 \\
Speech   & 146 \\
Social   & 142 \\
\midrule
\textbf{Total} & \textbf{600} \\
\bottomrule
\end{tabular}
\caption{Domain distribution of source--translation pairs.}
\label{tab:dataset-domains}
\end{table}

\paragraph{Translation systems.}
The 600 segments span 14 translation systems
(\Cref{tab:dataset-systems}) including frontier closed-source
LLMs (Claude-3.5/4, GPT-4/4.1, Gemini~2.5), production MT
systems (Dubformer, ONLINE-G, Yandex Translate), open-weight
LLMs (Llama3-70B, DeepSeek-V3), dedicated MT systems
(Unbabel-Tower70B), and human reference translations from the
WMT campaign.

\begin{table}[ht]
\centering
\small
\begin{tabular}{lr}
\toprule
\textbf{Translation system} & \textbf{Segments} \\
\midrule
Gemini~2.5             & 69 \\
Claude-3.5             & 55 \\
Dubformer              & 55 \\
GPT-4                  & 53 \\
ONLINE-G               & 53 \\
Unbabel-Tower70B       & 53 \\
Yandex Translate       & 52 \\
Reference (human)      & 52 \\
Hunyuan                & 33 \\
DeepSeek-V3            & 29 \\
Llama3-70B             & 27 \\
Claude~4               & 25 \\
Internal experimental  & 25 \\
GPT-4.1                & 19 \\
\midrule
\textbf{Total}         & \textbf{600} \\
\bottomrule
\end{tabular}
\caption{Translation system distribution.}
\label{tab:dataset-systems}
\end{table}

\paragraph{Annotation density.}
Each translation is annotated independently by three expert
linguists under the RATE protocol (Expert~$\times$~3). Combining
two annotations, translations contain on average $10.64$ error
spans per segment (\Cref{tab:dataset-severity}), with severity~2
(minor errors) most frequent followed by severity~3 and
severity~4 (major). Critical errors (severity~5) and trivial
errors (severity~1) together account for roughly $11\%$ of
annotated spans. The per-individual annotator rate is
approximately halved ($5.48$ spans per segment), consistent with
partial overlap across independent expert annotations.

\begin{table}[ht]
\centering
\small
\resizebox{\columnwidth}{!}{%
\begin{tabular}{lccc}
\toprule
\textbf{Severity} & \textbf{Combined} & \textbf{Annotator~A} & \textbf{Annotator~B} \\
\midrule
1 (trivial)   & 0.65  & 0.37 & 0.29 \\
2 (minor)     & 4.16  & 2.16 & 2.00 \\
3 (minor)     & 3.60  & 1.83 & 1.77 \\
4 (major)     & 1.71  & 0.85 & 0.86 \\
5 (critical)  & 0.53  & 0.28 & 0.25 \\
\midrule
\textbf{Total} & \textbf{10.64} & \textbf{5.48} & \textbf{5.22} \\
\bottomrule
\end{tabular}}
\caption{Mean error spans per segment by severity.
``Combined'' aggregates two independent expert annotations.
Per-annotator columns show individual outputs.}
\label{tab:dataset-severity}
\end{table}

\paragraph{Category distribution.}
\Cref{tab:dataset-categories} reports the distribution across
RATE error categories. Fluency errors dominate ($6.06$ spans per
segment, combined), followed by Mistranslation ($2.34$). All
other categories average below $1$ span per segment. This
long-tailed distribution motivates coverage-based few-shot
selection (\Cref{sec:rq2}): uniform random sampling would
oversample Fluency and Mistranslation while underrepresenting
rarer categories.

\begin{table}[ht]
\centering
\small
\begin{tabular}{lc}
\toprule
\textbf{Category} & \textbf{Spans / segment} \\
\midrule
Fluency           & 6.06 \\
Mistranslation    & 2.34 \\
Undertranslation  & 0.61 \\
Overtranslation   & 0.44 \\
Grammar           & 0.32 \\
Other             & 0.31 \\
Named Entity      & 0.30 \\
Style             & 0.29 \\
Omission          & 0.01 \\
\bottomrule
\end{tabular}
\caption{Mean error spans per segment by RATE category,
aggregated across the two expert gold annotations.}
\label{tab:dataset-categories}
\end{table}

\begin{table}[t]
\centering
\scriptsize
\setlength{\tabcolsep}{1.5pt}
\resizebox{\columnwidth}{!}{%
\begin{tabular}{l ccc ccc ccc}
\toprule
 & \multicolumn{3}{c}{$R$} & \multicolumn{3}{c}{$P$} &
   \multicolumn{3}{c}{$F_1$} \\
\cmidrule(lr){2-4}\cmidrule(lr){5-7}\cmidrule(lr){8-10}
\textbf{System} & MP & MPP & LLM & MP & MPP & LLM & MP & MPP & LLM \\
\midrule
Human\textsubscript{E}                       & 46.8 & 41.9 & 54.1 & 44.3 & 38.6 & 52.1 & 45.5 & 40.2 & 53.1 \\
Human\textsubscript{A-SbS}                   & 39.3 & 34.8 & 47.6 & 47.1 & 40.8 & 55.2 & 42.8 & 37.5 & 51.1 \\
\midrule
\textsc{Orbit-SC} GPT-5.4                    & 51.5 & 49.9 & 55.7 & 46.2 & 26.7 & 51.2 & 48.7 & 34.8 & 53.4 \\
\textsc{Orbit-SC} Claude                     & 47.8 & 45.9 & 57.1 & 46.9 & 30.6 & 51.4 & 47.3 & 36.8 & 54.1 \\
\textsc{Orbit-SC} Gemini                     & 45.0 & 42.8 & 47.6 & 51.2 & 36.9 & 56.7 & 47.9 & 39.6 & 51.7 \\
\textsc{Orbit-MC}                            & 60.3 & 58.2 & 66.2 & 45.8 & 28.6 & 49.1 & 52.1 & 38.4 & 56.4 \\
\bottomrule
\end{tabular}}
\caption{Span-level metrics under three span-matching
formulations: MP with $\tau{=}1$ (primary),
MPP~\citep{perrella2026spanlevel} (partial overlap with partial
credit), and an LLM reconciliation proxy using GPT-5.4. All
metrics are micro-averaged.
}
\label{tab:metric-robustness}
\end{table}

\begin{table*}[!t]
\centering
\small
\setlength{\tabcolsep}{5pt}
\begin{tabular}{l ccc cccc}
\toprule
 & \multicolumn{3}{c}{\textbf{Spans / segment}}
 & \multicolumn{4}{c}{\textbf{Severity-restricted metrics}} \\
\cmidrule(lr){2-4}\cmidrule(lr){5-8}
\textbf{Model} & Non-trivial & Major & Critical
& R\textsubscript{4+} & P\textsubscript{4+}
& $R_{4+}^\bullet$ & $P_{4+}^\bullet$ \\
\midrule
Human\textsubscript{E}        & 5.28 & 1.01 & 0.28 & 65.3 & 65.3 & 54.0 & 52.5 \\
\midrule
GPT-4.1                       & 5.67 & 0.93 & 0.28 & 72.7 & 61.7 & 53.1 & 56.1 \\
GPT-5.2                       & 4.26 & 0.90 & 0.27 & 64.8 & 67.5 & 45.4 & 59.1 \\
GPT-5.4                       & 5.57 & 1.55 & 0.49 & 74.7 & 63.5 & 61.3 & 55.7 \\
\midrule
Claude Sonnet 4.6             & 5.02 & 1.10 & 0.33 & 67.1 & 63.5 & 56.8 & 53.7 \\
Claude Opus 4.6               & 5.09 & 0.77 & 0.34 & 72.2 & 68.3 & 56.8 & 57.4 \\
\midrule
Gemini 3 Flash                & 3.96 & 0.66 & 0.23 & 62.2 & 66.8 & 47.6 & 60.7 \\
Gemini 3.1 Pro                & 4.39 & 1.09 & 0.24 & 67.8 & 66.9 & 56.8 & 61.0 \\
\midrule
Grok-4                        & 7.03 & 1.36 & 0.29 & 71.0 & 52.5 & 59.0 & 42.7 \\
DeepSeek-V3.2                 & 11.96 & 6.12 & 1.59 & 60.7 & 15.3 & 53.4 & 17.4 \\
DeepSeek-R1                   & 5.65 & 1.68 & 0.40 & 59.2 & 46.6 & 52.1 & 39.3 \\
Kimi K2                       & 3.49 & 0.70 & 0.18 & 39.4 & 50.5 & 32.6 & 39.2 \\
Kimi K2 Thinking              & 4.84 & 1.09 & 0.31 & 59.5 & 50.4 & 43.7 & 46.8 \\
Qwen-235B                     & 4.94 & 1.66 & 0.58 & 59.3 & 50.7 & 53.0 & 44.4 \\
\bottomrule
\end{tabular}
\caption{Extended model comparison using \textsc{Orbit-SC}
(RATE-AE\textsubscript{brief} with coverage-based $k{=}2$
demonstrations). Non-trivial, Major, and Critical report mean
predicted spans per segment at severity $\geq 2$, $\geq 4$, and
$=5$ respectively. $R_{4+}$ measures recovery of gold major
spans regardless of predicted severity, while $P_{4+}$ uses
predictions assigned severity $\geq 4$. The bullet variants use
predicted severity $\geq 3$, providing one-level tolerance around
the major threshold.}
\label{tab:rq3-extended}
\end{table*}

\paragraph{Source length and overall quality.}
The average source segment length is $377$ characters. The mean
RATE overall score (a derived 0--10 quality score) is $9.30$,
indicating that the dataset contains publication-grade
translations rather than failure-mode outputs. Mean annotated
span coverage per translation is $62$ characters, roughly $16\%$
of source length, with errors concentrated in localized regions
rather than spread diffusely.

\section{Artifact Use and Licenses}
\label{app:artifacts}

WMT24 General MT data is released by the WMT 
organizers under terms permitting research use. 
WMT23 MQM annotations~\citep{freitag2023results} 
are released under the Apache 2.0 license via the 
\texttt{google/wmt-mqm-human-evaluation} repository. 
WMT25 General MT data, including ESA 
annotations~\citep{lavie2025findings}, is publicly 
released by the WMT organizers via the 
\texttt{wmt-conference/wmt25-general-mt} repository 
for research purposes. xCOMET-XXL~\citep{guerreiro2024xcomet} 
is released under CC-BY-NC-SA-4.0. We use it for 
non-commercial research evaluation, consistent with 
its license terms. Closed-source LLM APIs (GPT, 
Claude, Gemini) were accessed in accordance with 
their respective terms of service.

Our use of all existing artifacts is consistent with
their intended research and non-commercial evaluation
purposes. The expert-annotated dataset, evaluation prompts, and
accompanying paper artifacts are available at
\url{https://huggingface.co/datasets/yandex/orbit-mt-eval}.

\section{Annotator Pool}
\label{app:annotators}

\paragraph{Expert annotators.}
Our Expert tier consists of professional translators who are
native Russian speakers with verified C1/C2 English proficiency
and substantial industry experience. A university degree in
linguistics or translation studies was preferred. Candidates
passed a multi-stage selection process: a post-editing test that
required context analysis and source fact-checking, followed by
an interview. After selection, experts received structured
training on the RATE protocol --- error categorization, severity
calibration, and the Accuracy/Fluency scoring scales --- with
concrete examples for each category.

\paragraph{Assessor.}
Our Assessor tier consists of in-house crowd-level evaluators:
native Russian speakers with verified C1 English proficiency
(screened via a standardized proficiency test). We do
\emph{not} require linguistics or translation degrees or
professional translation experience. Assessors received the
same RATE-protocol training materials as experts but were not
subject to the post-editing selection test.

\paragraph{Pool size and workload.}
The pool comprises \textbf{35}\,expert translators and
\textbf{82}\,assessors. Each segment is annotated independently
by three experts from the pool in implicit side-by-side mode
(Expert\,$\times$\,3), and by assessors in two setups: implicit
side-by-side (Human\textsubscript{A-SbS}) and point-wise
(Human\textsubscript{A-PW}). For evaluation we hold two expert
annotations as gold and score the third against each of them
and average, giving the Human\textsubscript{E} agreement
reported in Table~\ref{tab:main}
(Section~\ref{sec:baselines}). Average workload per annotation
pass was \textbf{17}\,segments per expert and \textbf{22}\,segments
per assessor.
Median annotation times were \textbf{14}\,minutes per segment
for experts and \textbf{5}\,minutes for assessors, compared
to the $\sim$\,68\,seconds per segment that
\citet{kocmi2025findings} report for English--Russian WMT25 ESA
annotation.

\paragraph{Training and dispute resolution.}
Before annotation began, all annotators participated in a
calibration session in which a small set of pre-annotated
examples was reviewed and edge cases were discussed
collaboratively. For the few-shot pool used in
Section~\ref{sec:rq2}, annotators worked in overlapping
groups on subsets of segments and reconciled disagreements
through discussion to produce consensus annotations more
reliable than any single-annotator output.

\section{Methods Not Included in Main Comparison}
\label{app:methods-not-included}
We do not include several recent MT evaluation approaches in
our primary comparison due to differences in evaluation setup
and workflow design. M-MAD~\citep{chen2024mmad} explores
debate-style multi-agent evaluation with substantially different
orchestration strategies and computational budgets. We draw on
its decomposition idea in our multi-call ablations
(\Cref{sec:rq4}) but do not reproduce the full multi-agent
system. EAPrompt~\citep{lu2024error} produces error analyses
through iterative prompting. A controlled re-evaluation
by~\citet{chen2024mmad} reports lower span-level $F_1$ than
GEMBA-MQM, which is already included in our comparison.
MQM-APE~\citep{lu2024mqmape}
relies on automatic post-editing as a validation step, requiring
multiple sequential LLM passes per segment and following a
substantially different multi-stage workflow. More broadly, our
goal is not to exhaustively benchmark every existing MT
evaluation system, but to study evaluator design principles
under a controlled experimental setup.

\section{Sensitivity to Span Matching}
\label{app:metric-robustness}

\citet{perrella2026spanlevel} recommend MPP with micro-averaging
to penalize imprecise span boundaries and avoid the preference of
MP for longer spans. Our target places more emphasis on
error-identity agreement, so we retain MP with $\tau{=}1$ as the
primary formulation and compare it with MPP and an LLM-based
reconciliation proxy. All metrics in this section are
micro-averaged.

\subsection{Formal Definitions}
\label{app:metric-defs}

We compare three matching formulations.

\paragraph{MP with $\tau{=}1$ (primary).}
Predicted error spans are matched to gold annotations via
bipartite matching that maximizes the number of matched pairs.
A predicted--gold pair is eligible for matching iff their
character overlap is $\geq 1$. Each matched pair contributes a
credit of $1.0$ to both precision (over predicted spans) and
recall (over gold spans).

\paragraph{MPP~\citep{perrella2026spanlevel}.}
Eligible predicted--gold pairs must overlap by at least one
character. Among eligible one-to-one matchings, MPP selects the
matching that maximizes the resulting MPP $F_1$. Credit is then
proportional to overlap. For
a matched pair with $k$ overlapping characters, precision credit
(over the predicted span of length $\ell_p$) is $k/\ell_p$, and
recall credit (over the gold span of length $\ell_g$) is
$k/\ell_g$.

\paragraph{LLM-based reconciliation proxy.}
For each translation, we present the predicted spans and gold
annotations to GPT-5.4 and instruct it to match annotations that
describe the same error, allowing character boundaries to differ.
Each matched pair contributes a credit of $1.0$. We treat this as
an exploratory reconciliation proxy rather than human
adjudication.

\begin{table*}[!t]
\centering
\setcounter{subtable}{0}
\begin{subtable}[t]{0.495\textwidth}
\centering
\scriptsize
\setlength{\tabcolsep}{1.6pt}
\renewcommand{\arraystretch}{0.96}
\begin{tabular}{@{}lccccc ccc@{}}
\toprule
 & \multicolumn{5}{c}{\textbf{Span-level}} & \multicolumn{3}{c}{\textbf{Ranking}} \\
\cmidrule(lr){2-6}\cmidrule(lr){7-9}
\textbf{Configuration} & R & P & F1 & R\textsubscript{4+} & P\textsubscript{4+} & MQM & Acc & FL \\
\midrule
GEMBA-MQM                         & 21.9 & 43.8 & 29.2 & 42.6 & 55.0 & 53.0 & -- & -- \\
GEMBA-ESA                         & 24.7 & 43.0 & 31.4 & 46.7 & 58.3 & 53.7 & -- & -- \\
MQM-AE                            & 24.5 & 48.0 & 32.4 & 46.7 & 63.7 & 54.0 & -- & -- \\
ESA-AE                            & 26.7 & 51.3 & 35.1 & 50.4 & 62.7 & 57.3 & -- & -- \\
UMB-v5                            & 23.5 & 42.5 & 30.3 & 46.1 & 61.8 & 54.7 & -- & -- \\
\addlinespace[1pt]
RATE-AE\textsubscript{brief}     & 29.5 & 54.4 & 38.2 & 54.4 & 72.4 & 59.3 & 67.7 & 65.3 \\
RATE-AE\textsubscript{full}      & 26.6 & \textbf{55.8} & 36.0 & 52.2 & \textbf{78.3} & 61.0 & 68.7 & 64.3 \\
RATE-AE\textsubscript{full+ex}   & 29.2 & 52.4 & 37.5 & 54.4 & 74.2 & 56.3 & 69.0 & 62.7 \\
\addlinespace[1pt]
RATE-AE\textsubscript{full+ex} + coverage, $k{=}2$ & 45.1 & 49.4 & \textbf{47.2} & 70.3 & 68.5 & \textbf{65.7} & \textbf{71.0} & 70.0 \\
RATE-AE\textsubscript{full+ex} + coverage, $k{=}4$ & \textbf{48.4} & 46.4 & \textbf{47.4} & 72.1 & 65.5 & 63.7 & 65.0 & 68.0 \\
\textsc{Orbit-SC}                & 47.8 & 46.9 & \textbf{47.3} & \textbf{72.2} & 68.3 & \textbf{65.7} & 68.3 & \textbf{70.7} \\
\bottomrule
\end{tabular}
\caption{Claude Opus 4.6.}
\label{tab:backbone-replication-claude}
\end{subtable}\hfill
\begin{subtable}[t]{0.495\textwidth}
\centering
\scriptsize
\setlength{\tabcolsep}{1.6pt}
\renewcommand{\arraystretch}{0.96}
\begin{tabular}{@{}lccccc ccc@{}}
\toprule
 & \multicolumn{5}{c}{\textbf{Span-level}} & \multicolumn{3}{c}{\textbf{Ranking}} \\
\cmidrule(lr){2-6}\cmidrule(lr){7-9}
\textbf{Configuration} & R & P & F1 & R\textsubscript{4+} & P\textsubscript{4+} & MQM & Acc & FL \\
\midrule
GEMBA-MQM                         & 20.3 & 45.4 & 28.1 & 40.2 & 60.5 & 59.3 & -- & -- \\
GEMBA-ESA                         & 16.3 & 47.2 & 24.2 & 36.1 & 57.9 & 52.3 & -- & -- \\
MQM-AE                            & 15.3 & 56.2 & 24.0 & 36.5 & 63.1 & 56.7 & -- & -- \\
ESA-AE                            & 19.5 & \textbf{60.1} & 29.4 & 42.0 & 69.4 & 60.0 & 57.7 & -- \\
UMB-v5                            & 20.6 & 47.9 & 28.8 & 43.7 & 68.6 & 56.3 & -- & -- \\
\addlinespace[1pt]
RATE-AE\textsubscript{brief}     & 21.7 & 54.9 & 31.1 & 45.1 & 73.5 & 58.0 & 53.7 & 61.7 \\
RATE-AE\textsubscript{full}      & 20.8 & 59.7 & 30.9 & 42.6 & \textbf{76.6} & 59.0 & 55.7 & 60.3 \\
RATE-AE\textsubscript{full+ex}   & 20.7 & 59.6 & 30.7 & 43.4 & 75.8 & 58.7 & 58.7 & 61.3 \\
\addlinespace[1pt]
RATE-AE\textsubscript{full+ex} + coverage, $k{=}2$ & 33.9 & 52.3 & 41.1 & 58.2 & 69.7 & 65.3 & \textbf{59.7} & 62.3 \\
RATE-AE\textsubscript{full+ex} + coverage, $k{=}4$ & 31.0 & 54.4 & 39.5 & 54.8 & 70.0 & 67.3 & 59.3 & 65.0 \\
\textsc{Orbit-SC}                & \textbf{38.9} & 49.1 & \textbf{43.4} & \textbf{62.2} & 66.8 & \textbf{68.0} & 59.3 & \textbf{67.0} \\
\bottomrule
\end{tabular}
\caption{Gemini 3 Flash.}
\label{tab:backbone-replication-gemini}
\end{subtable}
\caption{Matched design comparison across two backbones.
\textsc{Orbit-SC} uses
RATE-AE\textsubscript{brief} with curated coverage ($k{=}2$).}
\label{tab:backbone-replication}

\end{table*}

\subsection{Cross-Formulation Comparison}
\label{app:metric-table}

Table~\ref{tab:metric-robustness} reports all three
formulations across our main systems.

\subsection{Boundary Sensitivity under MPP}
\label{app:metric-boundaries}

The matching formulation affects exact system rankings and the
comparison between human and model agreement. Under MPP,
Human\textsubscript{E} drops by $4.9$ recall points and $5.7$
precision points. Across the three \textsc{Orbit-SC} backbones,
recall changes by only $1.6$--$2.2$ points, while precision falls
by $14.3$--$19.5$ points. \textsc{Orbit-MC} shows the same
asymmetry. This follows directly from MPP's partial-credit rule
when a predicted span extends beyond the gold boundaries.

Boundary fidelity is relevant for some downstream uses. However,
MPP can under-credit error-identity agreement when systems and
experts use different span granularities.

\subsection{LLM Reconciliation Proxy}
\label{app:metric-llm}

The LLM reconciliation proxy provides a complementary view. It
places Human\textsubscript{E} at $F_1{=}53.1$, the three
\textsc{Orbit-SC} configurations at $51.7$--$54.1$, and
\textsc{Orbit-MC} at $56.4$. These results support the main
error-identity comparison, while MPP shows that the exact
human--model ordering is sensitive to span boundaries.

GPT-5.4 is both the reconciler and one of the evaluated backbones,
so the proxy may share biases with the systems being judged. We
therefore use it as complementary evidence rather than as the
primary meta-evaluation method.

\subsection{Methodological Takeaway}
\label{app:metric-summary}

We retain MP with $\tau{=}1$ as primary because our main question
is whether an evaluator identifies the annotated error location.
MPP provides a boundary-sensitive view, while the LLM
reconciliation proxy emphasizes error identity. The three
formulations support the same broad conclusion that
\textsc{Orbit} is a strong evaluator, but exact system ordering
and the human--model comparison depend on the operational
definition of a span match.

\section{Model Comparison: Extended Results}
\label{app:rq3-extended}

\Cref{tab:rq3-extended} extends the model comparison from
\Cref{sec:rq3} with per-model span density (mean predicted
spans per segment, broken down by severity) and
severity-calibrated variants of the major-error metrics.

\paragraph{Severity-aware metrics.}
The main $R_{4+}$ metric measures recovery of gold spans with
severity $\geq 4$ regardless of predicted severity. The main
$P_{4+}$ metric measures precision for predictions assigned
severity $\geq 4$ against the same gold major-error spans. For
$R_{4+}^\bullet$ and $P_{4+}^\bullet$, only predictions assigned
severity $\geq 3$ are eligible for matching. These are
severity-aware variants with a one-level tolerance around the
major threshold. The gap between $R_{4+}$ and
$R_{4+}^\bullet$ therefore measures how often real major errors
are detected but assigned severity $\leq 2$.

\paragraph{Severity calibration is a separate axis of evaluator
behavior.}
As shown in \Cref{tab:rq3-extended}, the gap between $R_{4+}$
and $R_{4+}^\bullet$ reveals
substantial under-rating of major errors across both human and
LLM evaluators. Human\textsubscript{E} itself drops $11$ points
($65.3 \to 54.0$), confirming that severity assignment is
genuinely hard even for trained experts. LLM evaluators show
gaps of comparable magnitude: GPT-5.4 drops from $74.7$ to
$61.3$ ($-13.4$), Claude Opus~4.6 from $72.2$ to $56.8$
($-15.4$), and Gemini~3.1 Pro from $67.8$ to $56.8$ ($-11.0$).
This indicates that detection and severity assignment are
partially independent capabilities, and that the difficulty of
severity calibration is not specific to LLM evaluators ---
trained experts exhibit the same pattern. We adopt the
detection-oriented $R_{4+}/P_{4+}$ as primary metrics in the
main paper because they isolate the comparatively more
fundamental question of whether the evaluator finds the right
spans, but report severity-calibrated variants here as a
complementary diagnostic.

\paragraph{Span density patterns.}
The mean spans-per-segment columns clarify the recall--precision
trade-offs visible in the span-level metrics
(\Cref{tab:rq3}). Backbones with strong precision (Gemini~3.1
Pro, Claude Opus~4.6) predict at or below the expert non-trivial
rate ($4.39$ and $5.09$ vs.\ $5.28$), while recall-oriented
backbones (GPT-5.4, Grok-4) predict above it but with
major-severity counts that remain plausible ($1.55$ and $1.36$
vs.\ $1.01$). DeepSeek-V3.2 is the clear outlier: $11.96$
non-trivial spans and $6.12$ major errors per segment ---
roughly $6\times$ the expert major-error rate --- explaining
its severe precision collapse ($P_{4+} = 15.3$). Kimi-K2 sits
at the opposite extreme, under-predicting both non-trivial
($3.49$) and major ($0.70$) spans, consistent with its low
overall recall ($R = 22.0$).

\section{Cross-Backbone Design Replication}
\label{app:backbone-replication}

We evaluate the same 11 configurations with Claude Opus~4.6 and
Gemini~3 Flash using the final metric pipeline. This matched set
covers published baselines, RATE instruction variants,
coverage-based demonstrations, and the final \textsc{Orbit-SC}
configuration. The full comparison is reported in
\Cref{tab:backbone-replication}.

On Gemini, \textsc{Orbit-SC} is strongest on span $F_1$ and
$R_{4+}$. On Claude, it has the strongest $R_{4+}$ and is within
$0.1$ $F_1$ of the strongest configuration. RATE-based variants
also outperform the protocol baselines on $F_1$ for both backbones,
supporting transfer of the overall design while leaving individual
instruction effects backbone-dependent.

\section{Human--LLM Collaborative Annotation}
\label{sec:hmc}

We conduct a preliminary study of how expert annotators interact
with reports generated by the \textsc{Orbit-MC} multi-call
configuration. On a held-out set of 600 English--Russian
segments (disjoint from the design study data), we compare two
expert groups. Human\textsubscript{E1} first produced independent
annotations and then re-annotated the same segments after reviewing
\textsc{Orbit-MC} reports (post-hint, within-subject design).
Human\textsubscript{E2} received the reports before its initial
annotation (pre-hint).

In aggregate, access to \textsc{Orbit-MC} reports increases
annotation density and shifts annotators toward lower overall
quality scores (\Cref{tab:hmc}). Annotators do not adopt reports
uncritically: post-hint annotations retain fewer spans than the
raw report contains, and slider scores shift toward but do not
match \textsc{Orbit-MC}'s more severe assessments, indicating
partial rather than wholesale integration of report signals.
These results are exploratory --- the study measures annotation
revision rather than ground-truth correctness.

\begin{table}[ht]
\centering
\resizebox{\columnwidth}{!}{%
\setlength{\tabcolsep}{4pt}
\begin{tabular}{lccccc}
\toprule
 & \multicolumn{3}{c}{\textbf{Spans / segment}} & \multicolumn{2}{c}{\textbf{Mean scores}} \\
\cmidrule(lr){2-4}\cmidrule(lr){5-6}
\textbf{Annotator} & All & Major & Crit. & Acc & FL \\
\midrule
Human\textsubscript{E1} (baseline)           & 5.60 & 1.00 & 0.28 & 76.6 & 74.8 \\
Human\textsubscript{E1} + post-hint          & 7.04 & 1.11 & 0.30 & 72.3 & 72.1 \\
\hspace{0.5em}$\Delta$ within-subject        & +1.44 & +0.11 & +0.02 & $-$4.3 & $-$2.7 \\
\midrule
Human\textsubscript{E2} + pre-hint           & 6.00 & 1.03 & 0.26 & 74.1 & 73.0 \\
\midrule
\textsc{Orbit-MC} (raw hint)    & 8.09 & 0.78 & 0.24 & 60.2 & 59.7 \\
\bottomrule
\end{tabular}}
\caption{\small Effect of \textsc{Orbit-MC} hints on expert
annotation (held-out 600-segment set).}
\label{tab:hmc}
\end{table}

\end{document}